\documentclass[runningheads,dvipsnames]{llncs}
\pdfoutput=1

\usepackage[numbers,sort]{natbib}

\usepackage{graphicx}

\usepackage[nosumlimits]{amsmath}
\usepackage{amssymb}
\usepackage{booktabs}
\usepackage{multirow}

\usepackage[accsupp]{axessibility}

\usepackage{comment}

\usepackage{array}
\usepackage[inline]{enumitem}
\usepackage{stmaryrd}

\usepackage{algpseudocode}
\usepackage{algorithm}

\usepackage{caption}
\usepackage{color}

\usepackage{xspace}
\usepackage[inline]{enumitem}

\usepackage{bm}
\usepackage{caption}
\usepackage{subcaption}
\usepackage{balance}

\usepackage{appendix}
\usepackage[normalem]{ulem} %

\usepackage{wrapfig}
\usepackage{etoolbox}


\makeatletter \renewcommand\@biblabel[1]{#1.} \makeatother

\def\ECCVSubNumber{4927}

\newcommand{\myarraystretch}[0]{1.2}

\newcommand{\ourTitle}{Learning to Fit Morphable Models}

\newcommand{\supmat}{\textbf{\mbox{\textcolor{black}{Sup.~Mat.}}}\xspace}

\newcommand{\supmatlong}{\textbf{\mbox{\textcolor{black}{Supplementary~Material}}}\xspace}

\newcommand{\qheading}[1]{\noindent\textbf{#1}}

\newcommand{\etal}{et al.\xspace}
\newcommand{\ie}{i.e.\xspace}
\newcommand{\eg}{e.g.\xspace}

\newcommand{\wrt}[0]{\mbox{w.r.t.}\xspace}
\newcommand{\na}[0]{\mbox{n/a}\xspace}

\newcommand{\dof}[0]{\mbox{DOF}\xspace}

\newcommand{\hmd}[0]{\mbox{HMD}\xspace}

\newcommand{\distance}[1]{\llbracket#1\rrbracket}
\newcommand{\rnorm}[1]{\lVert#1\rVert_\psi}
\newcommand{\cardinality}[1]{\lvert#1\rvert}
\newcommand{\normabs}[1]{\lVert#1\rVert_1}
\newcommand{\normmse}[1]{\lVert#1\rVert_2}

\newcommand{\concat}[2]{[#1, #2]}

\newcommand{\amass}{\mbox{AMASS}\xspace}
\newcommand{\cmu}{\mbox{CMU}\xspace}
\newcommand{\kit}{\mbox{KIT}\xspace}
\newcommand{\mpihdm}{\mbox{MPI\_HDM05}\xspace}

\newcommand{\sixD}{6D\xspace}

\renewcommand{\eg}[0]{\mbox{e.g.}\xspace}
\renewcommand{\dof}[0]{\mbox{DoF}\xspace}
\newcommand{\mano}[0]{\mbox{MANO}\xspace}

\newcommand{\lgd}[0]{\mbox{LGD}\xspace}
\newcommand{\smpl}[0]{\mbox{SMPL}\xspace}
\newcommand{\smplh}[0]{\mbox{SMPL+H}\xspace}

\newcommand{\smplify}[0]{\mbox{SMPLify}\xspace}

\newcommand{\twoD}[0]{\mbox{2D}\xspace}
\newcommand{\threeD}[0]{\mbox{3D}\xspace}
\newcommand{\gru}[0]{\mbox{GRU}\xspace}
\newcommand{\grulong}[0]{Gated Recurrent Unit}

\newcommand{\threedpw}[0]{\mbox{3DPW}\xspace}

\newcommand{\optimname}[0]{\mbox{L-BFGS}\xspace}

\newcommand{\ground}[0]{\text{gnd}}

\newcommand{\vtovlong}[0]{\mbox{Vertex-to-vertex}\xspace}
\newcommand{\vtov}[0]{\mbox{V2V}\xspace}
\newcommand{\mplpe}[0]{\mbox{LdmkErr}\xspace}

\newcommand{\mpjpe}[0]{\mbox{JntErr}\xspace}
\newcommand{\pampjpe}[0]{\mbox{PA-MPJPE}\xspace}
\newcommand{\groundmetric}[0]{GrPe.\xspace}

\newcommand{\gt}[1]{\tilde{#1}}

\newcommand{\numjoints}[0]{\mathtt{J}}

\newcommand{\initregressor}[0]{\Phi}
\newcommand{\numverts}[0]{\mathtt{V}}
\newcommand{\params}[0]{\bm{\Theta}}
\newcommand{\intrinsics}[0]{\mathrm{K}}
\newcommand{\rotation}[0]{R}
\newcommand{\grad}[0]{\bm{g}}
\newcommand{\residuals}[0]{\mathcal{R}}
\newcommand{\data}[0]{D}
\newcommand{\optim}[0]{\mathcal{O}}
\newcommand{\damping}[0]{\bm{\lambda}}
\newcommand{\lr}[0]{\bm{\gamma}}
\newcommand{\internalloss}[0]{\loss^D}

\newcommand{\loss}[0]{\mathcal{L}}
\newcommand{\gravityloss}[0]{\loss^{\mathcal{G}}}
\newcommand{\up}[0]{\bm{\mathrm{u}}}
\newcommand{\prior}[0]{\text{prior}}
\newcommand{\poseprior}[0]{\loss^{\pose}_{\prior}}
\newcommand{\posepriorgmm}[0]{\loss^{\pose}_{\text{GMM}}}
\newcommand{\encoder}[0]{\mathcal{E}}
\newcommand{\posepriorvae}[0]{\loss^{\pose}_{\text{VAE}}}
\newcommand{\transform}[0]{T}
\newcommand{\temporal}[0]{\mathrm{T}}

\newcommand{\resnet}[0]{\mbox{ResNet}\xspace}
\newcommand{\relu}[0]{\mbox{ReLU}\xspace}

\newcommand{\update}[0]{u}
\newcommand{\pose}[0]{\bm{\theta}}
\newcommand{\shape}[0]{\bm{\beta}}
\newcommand{\expression}[0]{\bm{\psi}}
\newcommand{\transl}[0]{\bm{t}}
\newcommand{\sethree}{SE(3)}
\newcommand{\reals}{\mathbb{R}}
\newcommand{\mesh}{M}

\newcommand{\camera}{\Pi}
\newcommand{\cameraortho}{\Pi_o}
\newcommand{\camerapersp}{\Pi_p}

\newcommand{\camerascale}{s}
\newcommand{\cameratransl}{\bm{t}}
\newcommand{\joints}{\mathcal{J}}
\newcommand{\jointstwod}{\mathtt{j}}
\newcommand{\landmarks}{\mathcal{P}}
\newcommand{\numlandmarks}{\mathtt{P}}
\newcommand{\landmarkstwod}{p}

\newcommand{\edges}{\mathcal{E}}

\newcommand{\head}{\mathtt{H}}

\newcommand{\leftwrist}{\mathtt{L}}
\newcommand{\rightwrist}{\mathtt{R}}
\newcommand{\visibility}{v}

\newcommand{\mocap}{\mbox{MoCap}\xspace}

\newcommand{\ourrefcolor}{\textcolor{black}}
\newcommand{\eq}[1]{\mbox{\ourrefcolor{{Eq.}}~\ref{#1}}}
\newcommand{\alg}[1]{\mbox{\ourrefcolor{{Alg.}}~\ref{#1}}}
\newcommand{\tab}[1]{\mbox{\ourrefcolor{{Tab.}}~\ref{#1}}}
\newcommand{\fig}[1]{\mbox{\ourrefcolor{{Fig.}}~\ref{#1}}}

\newcommand{\Alg}[1]{\mbox{\ourrefcolor{{Algorithm}}~\ref{#1}}}

\newcommand{\Tab}[1]{\mbox{\ourrefcolor{{Table}}~\ref{#1}}}
\newcommand{\Fig}[1]{\mbox{\ourrefcolor{{Figure}}~\ref{#1}}}
\newcommand{\Sect}[1]{\mbox{\ourrefcolor{{Section}}~\ref{#1}}}

\usepackage{pifont}
\newcommand{\cmark}{\color{green}\ding{51}}
\newcommand{\xmark}{\color{red}\ding{55}}

\newcommand{\adam}[0]{\mbox{Adam}\xspace}
\newcommand{\adagrad}[0]{\mbox{ADAGRAD}\xspace}

\newcommand{\fullbody}[0]{\mbox{FB}\xspace}
\newcommand{\headshort}[0]{\mbox{H}\xspace}
\newcommand{\leftrighthands}[0]{\mbox{L / R}\xspace}

\newcommand{\lmlong}[0]{Levenberg–Marquardt\xspace}
\newcommand{\lm}[0]{\mbox{LM}\xspace}

\newcommand{\gd}[0]{\mbox{GD}\xspace}
\newcommand{\gdlong}[0]{Gradient Descent\xspace}
\newcommand{\gn}[0]{\mbox{GN}\xspace}
\newcommand{\gnlong}[0]{Gauss-Newton\xspace}

\newcommand{\powellong}[0]{Powell's dog leg method\xspace}
\newcommand{\powell}[0]{PDL}

\usepackage[dvipsnames]{xcolor}
\newcommand{\cameraready}[1]{#1\xspace}

\newcommand\blfootnote[1]{%
  \begingroup
  \renewcommand\thefootnote{}\footnote{#1}%
  \addtocounter{footnote}{-1}%
  \endgroup
}

\usepackage[capitalize]{cleveref}
\newcommand{\colorRef}[1]{\textcolor{black}{#1}} %
\crefname{figure}{\colorRef{Fig.}}{\colorRef{Figs.}}
\Crefname{figure}{\colorRef{Figure}}{\colorRef{Figures}}
\crefname{algorithm}{\colorRef{Alg.}}{\colorRef{Algs}}
\Crefname{algorithm}{\colorRef{Algorithm}}{\colorRef{Algorithms}}
\crefname{section}{Sec.}{Secs.}
\Crefname{section}{Section}{Sections}
\Crefname{table}{Table}{Tables}
\crefname{table}{Tab.}{Tabs.}

\begin{document}
\pagestyle{headings}
\mainmatter

\title{\ourTitle}
\titlerunning{\ourTitle}
\author{%
Vasileios Choutas\textsuperscript{1, 2, $\dagger$}%
\,
Federica Bogo\textsuperscript{2,\textsuperscript{*}}%
\,
Jingjing Shen\textsuperscript{2}
\,
Julien Valentin\textsuperscript{2}
}   

\institute{%
\textsuperscript{1}Max Planck Institute for Intelligent Systems, T{\"u}bingen, Germany,
\textsuperscript{2}Microsoft \\
{\tt\footnotesize
vchoutas@tue.mpg.de,
 fbogo@fb.com, \\ \{jinshen, valentin.julien\}@microsoft.com
}
\blfootnote{$^{\dagger}$Work performed at Microsoft. {*} Now at Meta Reality Labs Research.}
}
\authorrunning{Choutas \etal}

\maketitle

\begin{abstract}

Fitting parametric models of human bodies, hands or faces to sparse input signals in an accurate, robust, and fast manner has the promise of significantly improving immersion in AR and VR scenarios.
A common first step in systems that tackle these problems
is to regress the parameters of the parametric model directly from the input data.
This approach is fast, robust, and is a good starting
point for an iterative minimization algorithm.
The latter searches for the minimum of an energy function, typically composed of a data term and 
priors that encode our knowledge about the problem's structure.
While this is undoubtedly a very successful recipe,
priors are often hand defined heuristics and finding the right balance
between the different terms to achieve high quality results is a non-trivial task.
Furthermore, converting and optimizing these systems to run in a performant way requires custom implementations that demand 
significant time investments from both engineers and domain experts.
In this work, we build upon recent advances in learned
optimization and propose an update rule inspired by
the classic Levenberg–Marquardt algorithm. 
We show the effectiveness of the proposed
neural optimizer on three problems, 
\threeD body estimation from a head-mounted device,
\threeD body estimation from sparse \twoD keypoints
and face surface estimation from dense \twoD landmarks.
Our method can easily be
applied to new model fitting problems and offers a competitive alternative to well-tuned 'traditional' model fitting pipelines,
both in terms of accuracy and speed. 

\end{abstract}

\begin{figure}[t]
    \begin{center}
        \includegraphics[trim=000mm 000mm 000mm 000mm, clip=true, width=1.00\linewidth]{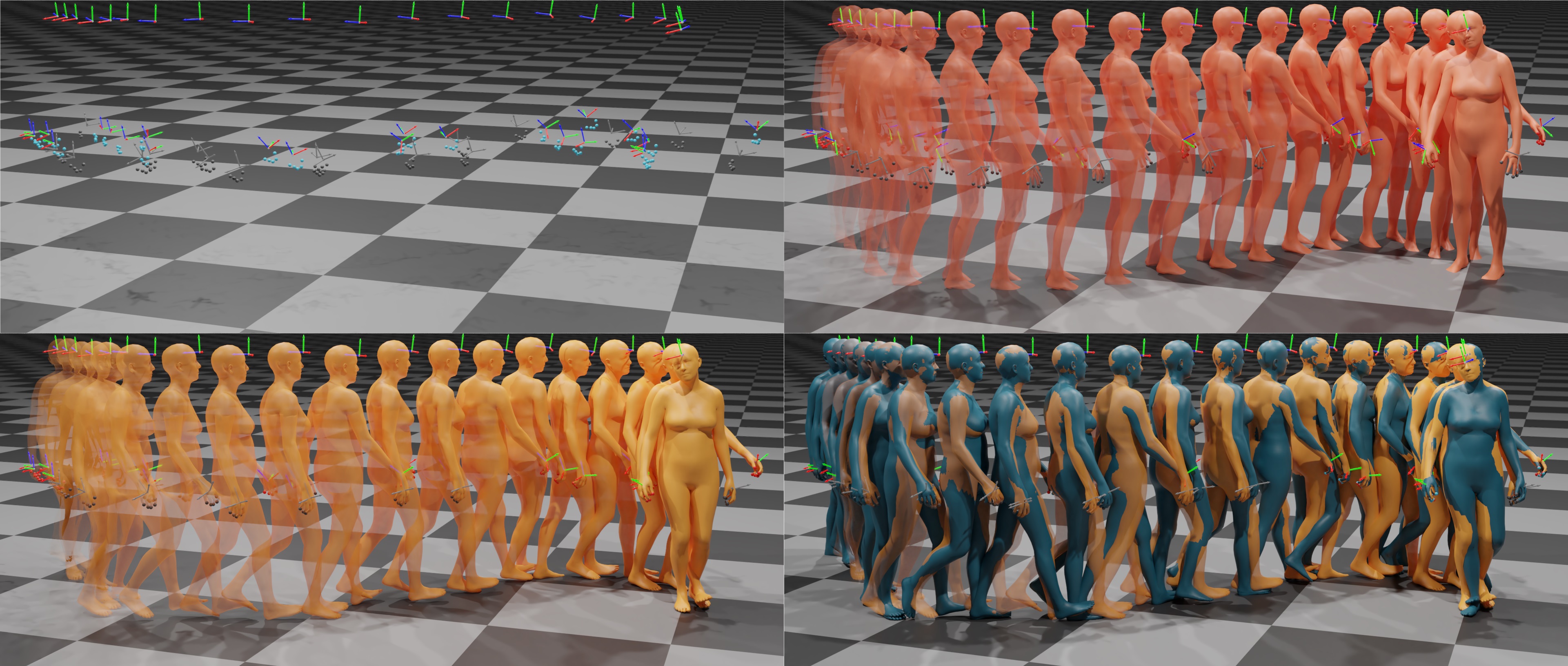}
    \end{center}
    \caption{%
        Top: Head and hand tracking signals from AR/VR devices (left) and
        the corresponding body model fit obtained from regression
        followed by iterative mathematical optimization.
        Bottom: Body model fit obtained from our learned optimizer (left),
        overlaid with the ground-truth (right).
        Learned optimizers are fast, able to tightly fit the input data
        and require significantly less manual labor to achieve this result.
        All results are estimated independently per-frame.
    }
    \label{fig:teaser}
\end{figure}
\section{Introduction}
\label{sec:intro}

Fitting parametric models \cite{3DMM_survey,mano,Pavlakos2019_smplifyx,Xu_2020_CVPR,Joo2018_adam,scape}
to noisy input data is one of the most 
common tasks in computer vision. 
Notable examples include
fitting 3D
body 
\cite{Bogo:ECCV:2016,Kocabas_2021_ICCV,Kolotouros_2019_ICCV,Pavlakos2019_smplifyx,Xiang2019,Choutas2020_expose,PIXIE:2021},
face \cite{3DMM_survey}, and
hands \cite{Baek_2019_CVPR,Boukhayma2019,hasson_2019_cvpr, PhongSurface2020}.

Direct regression using neural networks is
the de facto default tool to estimate model parameters from observations.
While the obtained predictions are robust and accurate to a large extent,
they often fail to tightly fit the observations \cite{pymaf2021}
and require large quantities of annotated data.
Classic optimization methods, \eg 
the \lmlong (\lm) algorithm \cite{levenberg1944method, marquardt1963algorithm}, can tightly fit the parametric
model to the data
by iteratively minimizing a hand-crafted energy function, but are prone to local minimas and require good starting points for fast convergence.
Hence, practitioners
combine these two approaches to benefit from their complementary strengths, 
initializing the model parameters from a regressor, followed
by %
energy minimization using a classic optimizer.

If we look one level deeper, optimization-based model fitting methods have another disadvantage of often requiring
hand-crafted energy functions that are difficult to define
and non-trivial to tune. Besides the data terms,
each fitting problem effectively requires the definition
of their own prior terms and regularization terms.
Besides the work
required to formulate these
terms
and train the priors, domain
experts needs to spend significant amounts of time to balance
the effect of each term. Since these priors
are often hand-defined or assumed to follow distributions that are tractable / easy to optimize, the resulting fitting energy usually contains biases that can limit the accuracy of the resulting fits.

To get the best of both regression using deep learning and classical numerical
optimization, we turn to the field of machine learning based continuous optimization \cite{schmidhuber1992learning, schmidhuber1993neural,Clark_2018_ECCV,NIPS2016_l2l,jie_eccv_2020,Zanfir_2021_CVPR}. 
Here, instead of updating the model parameters using a first or second order model fitter,
a network learns to iteratively update the parameters
that minimize the target loss,
with the added benefit of optimized ML back-ends for fast inference.
End-to-end network training removes the need for hand-crafted 
priors, since the model learns them directly from data.

Inspired by the properties of the popular
\lmlong and
\adam \cite{adam} algorithms,
our main contribution extends the system presented in
\cite{jie_eccv_2020} with an iterative machine learning
solver which 
\begin{enumerate*}[label=(\roman*)]
    \item keeps information from previous iterations,
    \item controls the learning rate of each variable independently and
    \item combines updates from gradient descent and from a network that is capable of swiftly reducing the fitting energy, 
    for robustness and convergence speed.
\end{enumerate*}
We evaluate our approach on different challenging scenarios: %
full-body tracking from head and hand inputs only,
\eg given by a device like the HoloLens 2,
body estimation from \twoD keypoints
and face tracking from \twoD landmarks, demonstrating both high quality results and versatility of the proposed framework. %

\section{Related Work}
\label{sec:related}

\qheading{Learning to optimize} 
\cite{schmidhuber1992learning, schmidhuber1993neural,NIPS2016_l2l}
is a field that,
casts optimization as a learning problem. The goal is 
to create models that learn to exploit the problem structure,
producing faster and more effective energy minimizers. In this way,
we can remove the need for hand-designed parameter update rules
and priors, since we can learn them directly from the data.
This approach has been used for image denoising and depth-from-stereo
estimation \cite{vogel_gcpr}, 
rigid motion estimation \cite{lv2019taking},
view synthesis \cite{flynn2019deepview},
joint estimation of motion and scene geometry \cite{Clark_2018_ECCV},
non-linear tomographic inversion problem with simulated data
\cite{adler2017solving}, face alignment \cite{Xiong_2013_CVPR}
and object reconstruction from a single image \cite{kokkinos2021to}.

\qheading{Parametric human model fitting:} 
The seminal work of Blanz and Vetter \cite{BlanzVetter1999}
introduced a parametric model of human faces and a user-assisted 
method to fit the model to images. Since then, the field has evolved
and produced better face models and faster, more accurate and more robust
estimation methods \cite{3DMM_survey}. With the introduction of
\smpl \cite{SMPL:2015}, the field of \threeD body pose and shape estimation
has been rapidly progressing. The community has
created large motion databases \cite{AMASS:ICCV:2019}
from motion capture data, as well as datasets,
both real and synthetic, with images and corresponding \threeD body ground-truth 
\cite{vonMarcard2018,agora,PROX:2019}. Thanks to these,
we can now train
neural network regressors that can reliably 
predict \smpl parameters from images \cite{joo2020eft,kanazawa_2018_cvpr,Kolotouros_2019_ICCV,pymaf2021,kolotouros2021prohmr,li2021hybrik} and
videos \cite{choi2020beyond,VIBE:CVPR:2020}.
With the introduction of expressive models
\cite{Joo2018_adam,Pavlakos2019_smplifyx,Xu_2020_CVPR}, the latest regression approaches
\cite{Choutas2020_expose,rong2021frankmocap,PIXIE:2021} can now predict the \threeD body, face and hands.
However, one common issue, present in all regression scenarios,
is the misalignment of the predictions and the
input data \cite{pymaf2021,Seeber:RealisticHands:2021}.
Thus, they often serve as the initial point for an
optimization-based method \cite{Bogo:ECCV:2016,Pavlakos2019_smplifyx,Xiang2019},
which refines the estimated parameters until some convergence
criterion is met.
This combination produces system that are effective, robust
and able to work in real-time and under challenging conditions
\cite{taylor_hands, PhongSurface2020, MuellerHands2019}.
These hybrid regression-optimization systems are also effective
pseudo annotators for in-the-wild images \cite{Kolotouros_2019_ICCV},
where standard capture technologies are not applicable.
However, formulating the correct energy terms
and finding the right balance between them is a challenging and time-consuming
task.
Furthermore, adapting the optimizer to run in real-time
is a non-trivial operation, even when using popular algorithms such as the 
\lmlong algorithm
\cite{levenberg1944method, marquardt1963algorithm, lm_cubic2005} which has a cubic complexity.
Thus, explicitly computing the Jacobian \cite{Clark_2018_ECCV,lv2019taking}
is often prohibitive in practice, 
either in terms of memory or runtime.
The most common and practical way to speedup the optimization is to utilize the sparsity of the problem or make certain assumptions to simplify it \cite{Fan_2021_ICCV}.
Learned optimizers promise to overcome these issues,
by learning the parametric model priors directly from the data
and taking more aggressive steps, thus converging in fewer iterations.
The effectiveness of these approaches has been demonstrated 
in different scenarios, such as fitting a body model \cite{SMPL:2015,Xu_2020_CVPR}
to images \cite{jie_eccv_2020,Zanfir_2021_CVPR} and 
videos \cite{yuan2021simpoe},
to sparse sensor data from electromagnetic sensors \cite{kaufmann2021pose}
and multi-body estimation from multi-view images \cite{dong2021shape}.

We propose a new
update rule, computed as a weighted combination of
the gradient descent step
and the network update \cite{jie_eccv_2020}, where their relative weights are
a function of the residuals. Many popular optimizers have an internal
memory, such as \adam's \cite{adam} running averages,
Clark \etal's \cite{Clark_2018_ECCV} and Neural Descent's
\cite{Zanfir_2021_CVPR} RNN. We adopt this insight, using an RNN to predict
the network update and the combination weights.
The network can choose
to follow either the gradient or the network direction more,
using both
current and past residual values. \looseness=-1

\noindent{\textbf{Estimating \threeD human pose from a head-mounted device}}
is a difficult problem,
due to self-occlusions caused by the position of the headset and the sparsity of the
input signals \cite{lower_body}. Yuan and Kitani 
\cite{Yuan_2018,yuan2019ego} cast this as a control problem, where a model
learns to produce target joint angles for a Proportional-Derivative (PD) controller.
Other methods \cite{tome2019xr,tome2020self} tackle this as a learning problem, where
a neural network learns to predict the \threeD pose from the cameras
mounted on the \hmd. Guzov \etal \cite{Guzov_2021_CVPR} use sensor data from
IMUs placed on the subject's body and combine them with camera self-localization.
They formulate an optimization problem with scene constraints, enabling
the capture of long-term motions that respect scene constraints,
such as foot contact with the ground. Finally, 
Dittadi \etal \cite{dittadi2021full-body} propose a likelihood model
that maps head and hand signals to full body poses.
In our work, we focus on this scenario and empirically show
that the proposed
optimizer rule is competitive, both with a classic optimization
baseline and a state-of-the-art likelihood model \cite{dittadi2021full-body}.

\section{Method}
\label{sec:method}

\subsection{Neural Fitter}
\label{subsec:neural fitter}

\lmlong (\lm) \cite{levenberg1944method, marquardt1963algorithm, lm_cubic2005}
and \powellong (\powell) \cite{Powell1970} 
are examples of
popular iterative optimization algorithms used in applications that fit either faces
or full human body models to observations. These techniques employ
the Gauss-Newton algorithm for both its convergence rate
approaching the quadratic regime and its computational efficiency,
enabling real-time model fitting applications, e.g. generative face
\cite{thies2016face2face,zollhofer2018state}
and hand \cite{taylor_hands, PhongSurface2020}
tracking.
For robustness, \lm and \powell~both combine the Gauss-Newton algorithm and gradient descent,
leading to implicit and explicit trust region being used when
calculating updates, respectively. In \lm, the relative contribution of the approximate
Hessian and the identity matrix is weighted by a single scalar that is changing
over iterations with its value carried over from one iteration to the next.
\cameraready{%
Given an optimization problem over a set of parameters $\Theta$,
\lm computes the parameter update $\Delta\params$ as the solution of the system
$(J^TJ + \lambda \text{diag}(J^TJ)) \Delta\params = J^T \residuals$,
where $J$ is the Jacobian and $\residuals$ are the current residual values.
}
It is interesting to note that several popular optimizers, including
\adagrad \cite{adagrad} and
\adam \cite{adam}, also carry over information about previous iteration(s),
in this case to help control the learning rate for each parameter.

Inspired by the success of these algorithms, we aim at constructing a novel neural optimizer that
\begin{enumerate*}[label=(\alph*)]
    \item is easily applicable to different fitting problems,
    \item can run at interactive rates without requiring significant efforts,
    \item does not require hand crafted priors.
    \item carries over information about previous iterations of the solve,
    \item controls the learning rate of each parameter independently,
    \item for robustness and convergence speed, combines updates
    from gradient descent and from a method capable
    of very quickly reducing the fitting energy.
\end{enumerate*}
Note that the Learned Gradient Descent (LGD) proposed in
\cite{jie_eccv_2020} achieves (a), (b), and (c),
but does not consider (d), (e), and (f). 
As demonstrated experimentally in Section~\ref{sec:experiments}, each of these additional properties leads to improved results compared to \cite{jie_eccv_2020}, and the best results are achieved when combined together.

\setlength\intextsep{0pt}
\begin{wrapfigure}{l}{0.5\textwidth} 
    \begin{minipage}{0.5\textwidth}
    \begin{algorithm}[H]
    \caption{Neural fitting}
    \label{alg:cap}
    \begin{algorithmic}
    \Require Input data $\data$
    \State $\params_0 = \initregressor\left(D\right)$
    \State $h_0 = \Phi_h\left(\data\right)$
    \While{not converged}
        \State $\Delta\params_n, h_n \gets f(\concat{\grad_{n-1}}{\params_{n-1}}, \data, h_{n-1}) $
        \State $\params_n \gets \params_{n-1} + \update\left(\Delta\params_n, \grad_{n-1}, \params_{n-1}\right)$
    \EndWhile
    \end{algorithmic}
    \label{alg:fitting}
    \end{algorithm}
\end{minipage}
\end{wrapfigure}

Our proposed neural fitter estimates the values
of the parameters $\params$ by
iteratively updating an initial
estimate $\params_0$,
see \Alg{alg:fitting}.
While the initial estimate $\params_0$ obtained
from a deep neural network $\initregressor$
might be sufficiently accurate
for some applications, we will show that
a careful construction of the update rule ($\update(.)$ in \alg{alg:fitting})
leads to significant improvements after only a few iterations.
It is important to note that
we do not focus
on building the best possible initializer $\initregressor$
for the fitting tasks at hand, which is the focus
of \eg VIBE \cite{VIBE:CVPR:2020} and SPIN \cite{Kolotouros_2019_ICCV}.
That being said, note that these regressors could be leveraged to
provide $\params_0$ from \alg{alg:fitting}.
$h_0$ and $h_n$ are the hidden states of the optimization process. 
At the $n$-th iteration in the loop of \alg{alg:fitting},
we use a neural network $f$ to predict $\Delta\params_n$,
and then apply the following update rule: 
\begin{align}
    & u(\Delta\params_{n}, \grad_{n-1}, \params_{n-1}) =
    \damping
    \Delta\params_n
    + (-\lr \grad_{n-1})
    \label{eq:line_search_like} \\
    & \damping, \lr =
     f_{\damping,\gamma}(\residuals(\params_{n-1}), \residuals(\params_{n-1} + \Delta\params_n))
    , \damping, \lr \in \reals^{\cardinality{\params}}
    \label{eq:gamma_damping}
\end{align}
Note that
\lgd \cite{jie_eccv_2020}
is a special case of \eq{eq:line_search_like}, with
$\lambda=1, \gamma=0$, and with no knowledge preserved across fitting iterations.
$\grad_n$ is the gradient of the target data
term \wrt to the problem parameters: $\grad_n=\nabla \internalloss$.

The proposed neural fitter satisfies the requirements (a), (b) and (c) in a similar fashion to LGD \cite{jie_eccv_2020}. In the following, we describe how the properties (d), (e), and (f) outlined earlier in this section are satisfied.

\qheading{(d): keeping track of past iterations.}
The functions $f, f_{\damping, \lr}$ are
implemented with a \grulong{ }(\gru) \cite{gru}. %
Unlike previous work, where the learned optimizer
only stores past parameter values and the total loss
\cite{Zanfir_2021_CVPR}, 
leveraging \gru{s} allows to learn an abstract representation of the knowledge
that is important to use and forget about the previous iteration(s),
and of the knowledge about the current iteration that should be
preserved.

\qheading{(e): independent learning rate.}
When fitting face or body models to data, the variables being optimized
over are of different nature. For instance, rotations might be expressed
in Euler angles while translation in meters. 
Since each of these parameter has a different scale
and / or unit, it is useful to have per-parameter step size values. Here, we propose to predict vectors $\lambda$ and $\gamma$ independently to scale the relative
contribution of $\Delta\params_n$ and $\grad_n$ to the update applied
to each entry of
$\params_n$. It is interesting to note that $f_\lambda$
having knowledge about the current value of residuals at $\params_n$
and the residual at $\params_n + \Delta\params_n$, effectively makes use of an estimate of the
step direction before setting a step size which is analogous to how line-search operates. 
Motivated by this observation we tried a few learned versions
of line search which yielded similar or inferior results to what
we propose here. The alternatives we tried are described in the \supmat.

\qheading{(f): combining gradient descent and network updates.}
\cameraready{%
\lm interpolates between \gdlong (\gd) and \gnlong (\gn)
using an iteration dependent scalar. \lm combines
the benefits of both approaches, namely fast convergence
near the minimum like \gn and large descent steps
away from the minimum like \gd. In this work, we replace
the \gn direction, which is often prohibitive to compute,
with a network-predicted update, described in \cref{eq:line_search_like}.
The neural optimizer should learn the optimal descent direction and the
relative weights to minimize the data term in as few steps
as possible.}
In the \supmat we provide alternative
combinations, \eg via convex combination,
which yielded inferior results in our experiments.

\subsection{Human Body Model and Fitting Tasks}
\label{subsec:human_model}

\renewcommand{\params}{\bm{\Theta}}

\ifdef{\captiondatasample}{}{
\newcommand{\captiondatasample}[0]{%
    Left to right: 1) Input
    6-\dof transformations
    $T_{\head}, T_{\leftwrist}, T_{\rightwrist}$
    and fingertip positions
    $P_{i=1, \ldots 5}^{\leftwrist}, P_{i=1, \ldots 5}^{\rightwrist}$,
    given by the head-mounted device, 2) ground-truth mesh,
    3) half-space visibility,
    everything behind the
    headset is not visible.
}
}

\setlength\intextsep{-1.5ex}
\begin{wrapfigure}[17]{l}{0.38\textwidth}
    \begin{center}
   \includegraphics[trim=260mm 000mm 240mm 000mm, clip=true, width=0.32\linewidth]{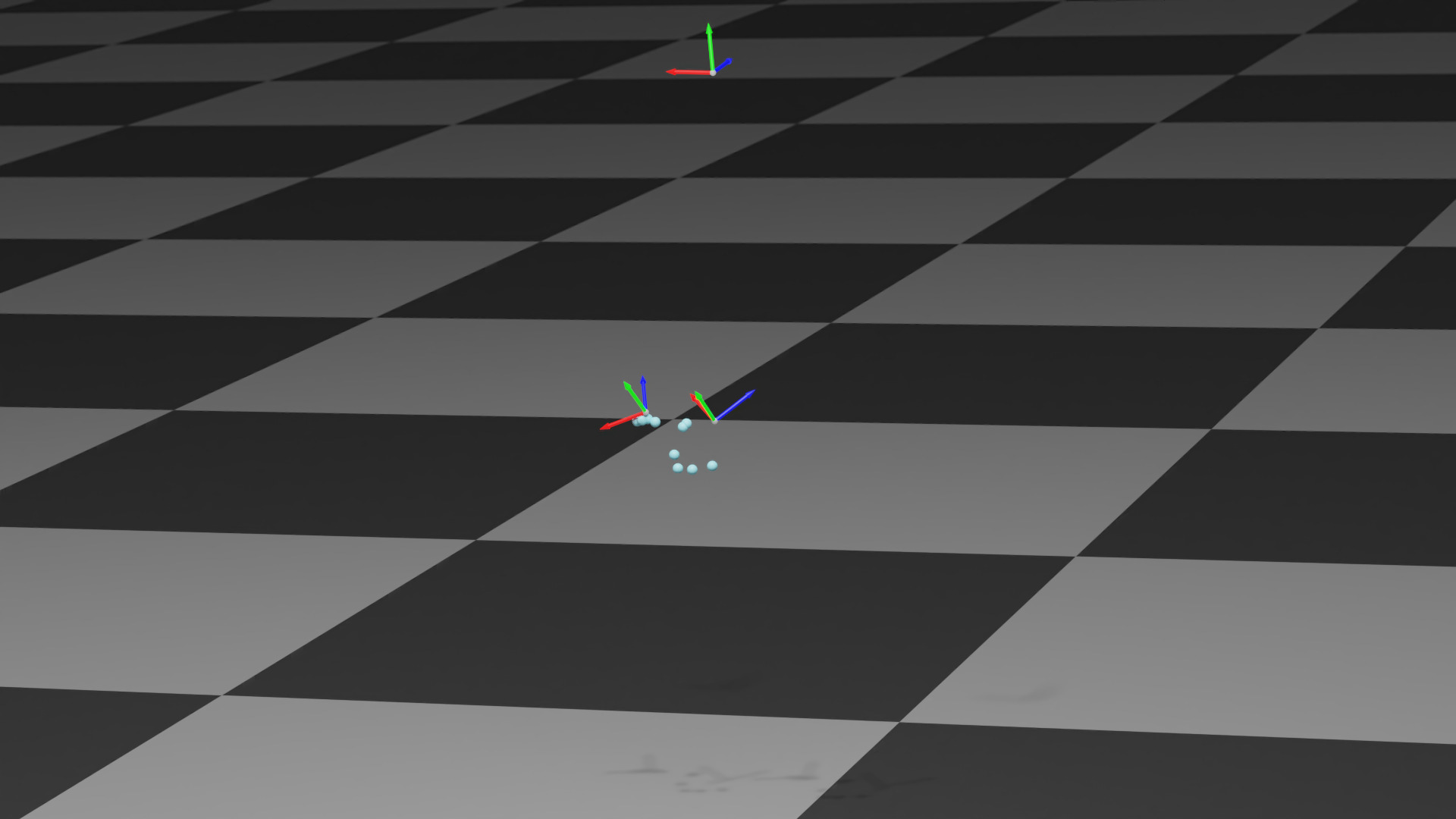}%
   \includegraphics[trim=260mm 000mm 240mm 000mm, clip=true, width=0.32\linewidth]{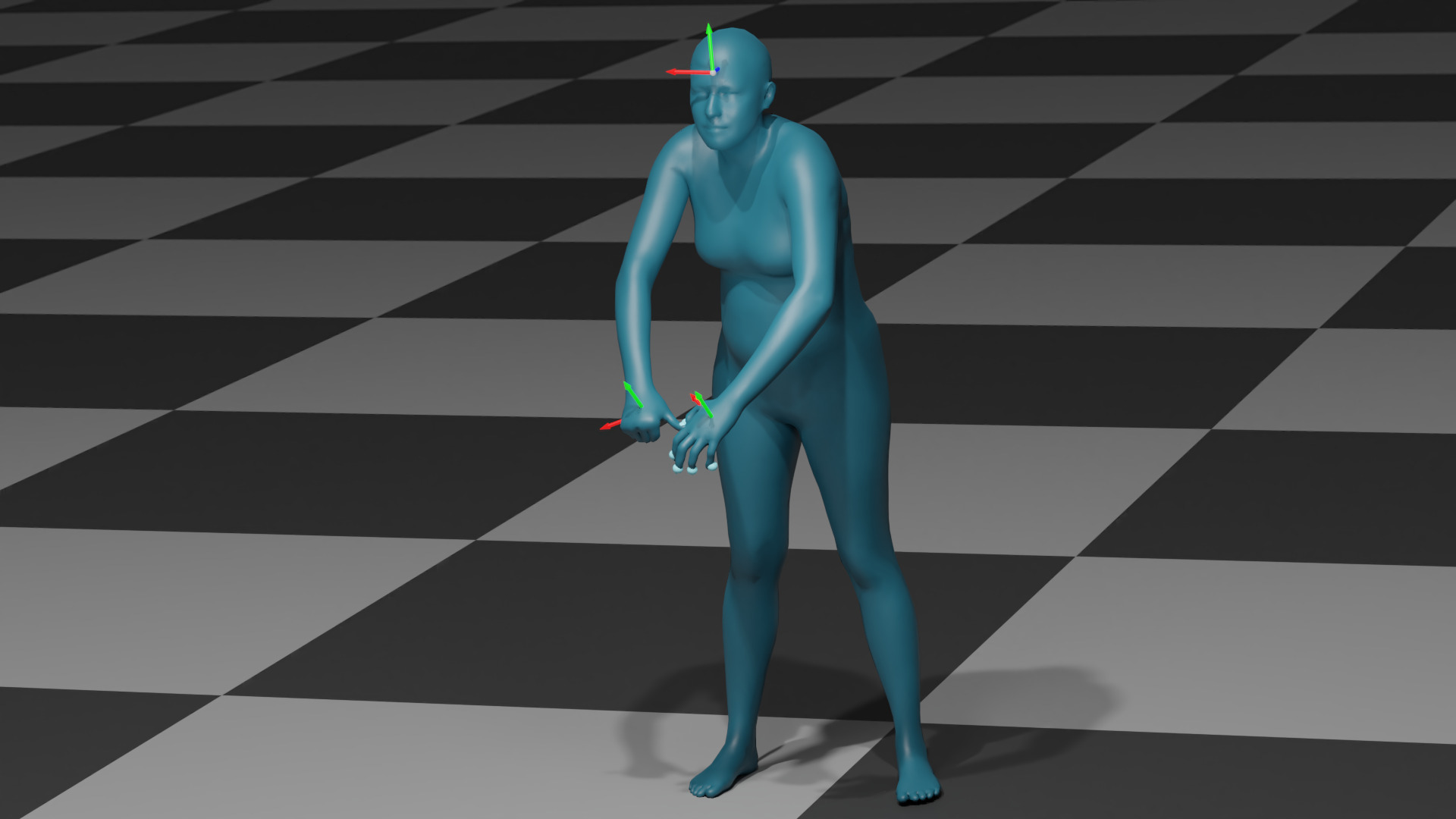}%
   \includegraphics[trim=260mm 000mm 240mm 000mm, clip=true, width=0.32\linewidth]{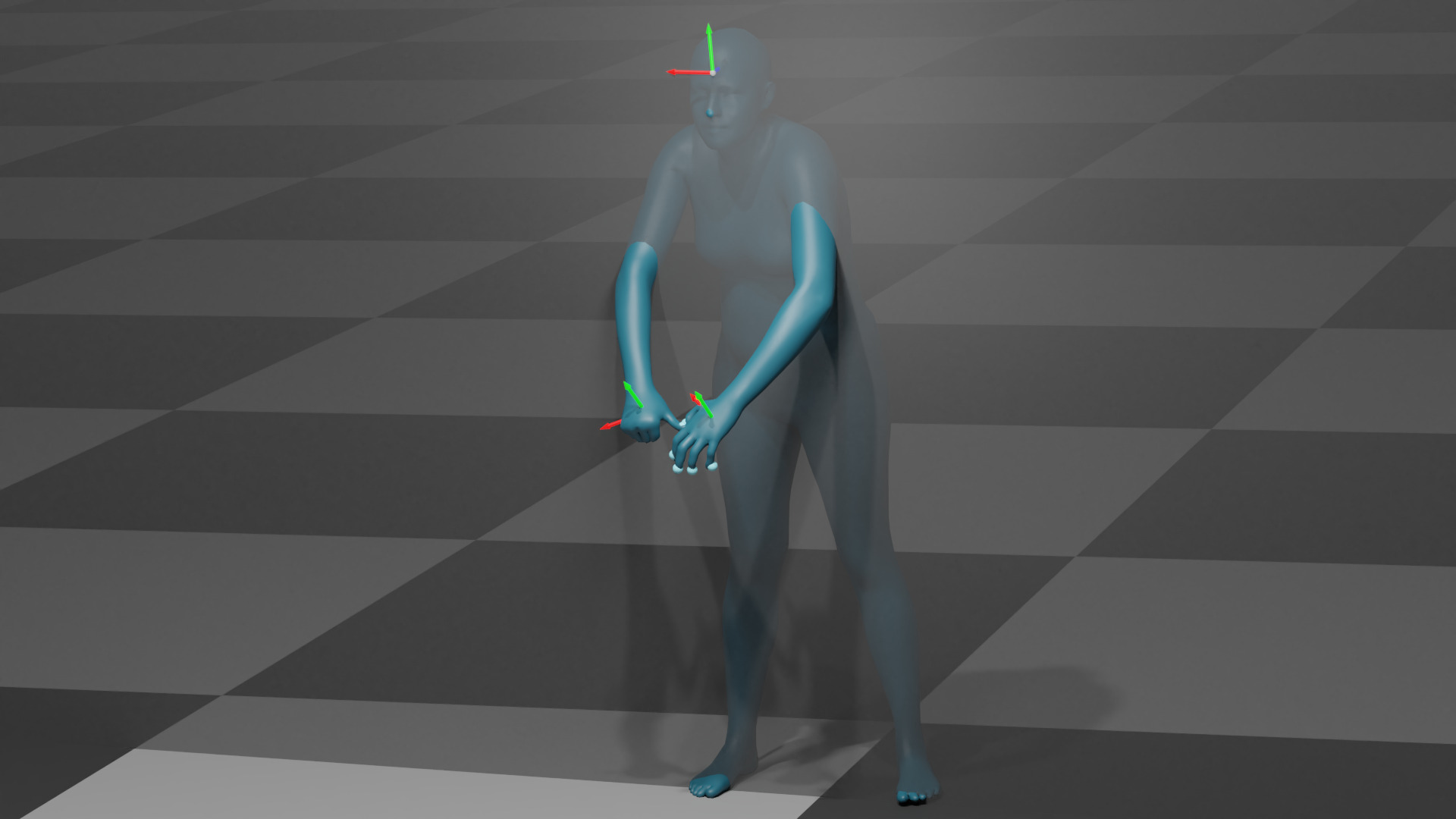}
    \caption{\captiondatasample}
    \label{fig:data_sample}
    \end{center}
\end{wrapfigure}

We represent the human body using \smpl \cite{SMPL:2015}/\smplh\cite{mano},
a differentiable function that computes 
mesh vertices $\mesh(\pose, \shape)
\in \reals^{\numverts \times 3}$, $\numverts=6890$,
from pose $\pose$
 and shape $\shape$, 
using standard linear blend skinning (LBS). 
The \threeD joints, $\mathcal{J}(\shape)$,
of a kinematic skeleton are computed from the shape parameters.
The pose parameters
$\pose \in \reals^{\numjoints \times \mathtt{D}  + 3}$
contain the parent-relative rotations of each joint and the root translation,
where $\mathtt{D}$ is the dimension of
the rotation representation and $\numjoints$ is the number 
of skeleton joints.
We represent rotations using
the \sixD rotation parameterization of
Zhou \etal \cite{Zhou_2019_CVPR},
thus $\theta \in \reals^{\numjoints \times 6 + 3}$.
The world transformation $T_j(\pose) \in \sethree$
of each joint $j$
is computed by following the transformations of its parents
in the kinematic tree:
$T_j(\pose) = T_{p(j)}(\pose) * T\left(\pose_j, \mathcal{J}_j(\shape)\right)$,
where $p(j)$ is the index of the parent of joint $j$
and $T\left(\pose_j, \mathcal{J}_j(\shape)\right)$ is the
rigid transformation of joint $j$ relative to its parent.
Variables with a 
\emph{hat} denote observed quantities.

We focus on two \threeD human body estimation problems:
1) fitting a body model \cite{SMPL:2015}
to \twoD keypoints and 2) inferring the body,  including
hand articulation \cite{mano}, from head and hand signals returned
by AR/VR devices, shown in \fig{fig:data_sample}.
The first is by now a standard problem in the Computer Vision community.
The second, which uses only head and hand signals in the AR/VR scenario, is a significantly harder task which requires strong priors, in particular to produce plausible results for the lower body and hands.
The design of such priors is not trivial,
\cameraready{requires expert knowledge
and a
significant investment of time.} \\
\qheading{\twoD keypoint fitting:}
We follow the setup of Song \etal \cite{jie_eccv_2020},
computing the projection of the \threeD \smpl joints $\joints$
with a weak-perspective camera $\camera$ with scale $\camerascale \in \reals{}$,
translation
$\cameratransl \in \reals^{2}$:
$\jointstwod = \cameraortho(\joints(\pose, \shape), \camerascale, \cameratransl) $.
Our goal is to
estimate \smpl and camera parameters $\params^{B} = \{\pose, \shape\}$, $\intrinsics^B = \{\camerascale, \cameratransl \}$, such that the
projected joints $\jointstwod$ match the detected keypoints 
$D^B = \{ \hat{\jointstwod}\}$, \eg from OpenPose \cite{openpose}.
\qheading{Fitting \smplh to AR/VR device signals:}
We make the following assumptions:
\begin{enumerate*}%
\item the device head tracking system provides a 6-\dof transformation
$\hat{T}^{\head}$, that contains the position and orientation of the \emph{headset} in the world coordinate frame.
\item the device hand tracking system gives us the orientation and position of the left and right wrist,
$\hat{T}^{\leftwrist}, \hat{T}^{\rightwrist} \in \sethree$,
and the positions of the fingertips
$\hat{P}_{1, \ldots, 5}^{\leftwrist}, 
\hat{P}_{1, \ldots, 5}^{\rightwrist} \in \reals^{3}$
in the world coordinate frame,
if and when they are in
the field of view (FOV) of the \hmd.
\end{enumerate*}
In order to estimate the \smplh parameters that best fit the 
above observations, we  
compute the estimated headset position and orientation from the
\smplh world transformations as
$T^{\head}(\params)=  T^{\text{\hmd}} T_{j_\head}(\params)$,
where $j_{\head}$ is the index of the head joint of \smplh.
$T^{\text{\hmd}}$ is a fixed transform from the \smplh head joint
to the headset, obtained from an offline calibration phase. \\
Visibility is represented by $\visibility_{\leftwrist},
\visibility_{\rightwrist} \in \{0, 1\}$
for the left and right hand respectively.
We examine two scenarios:
\begin{enumerate*}
    \item full visibility, where the hands are always visible, %
    \item half-space visibility, where only the area in front of the \hmd is visible.
\end{enumerate*}
Specifically, we transform
the points into the coordinate frame of the headset, using 
$T^{\head}$. All points with $z \geq 0$ are behind the headset
and thus invisible. \fig{fig:data_sample} right visualizes the plane that defines what is visible or not. \\
To sum up, the sensor data
are:
$D^{\text{\hmd}} = \{\hat{T}^{\head}, \hat{T}^{\leftwrist}, \hat{T}^{\rightwrist},
\hat{P}_{i=1, \ldots, 5}^{\leftwrist}, \hat{P}_{i=1, \ldots, 5}^{\rightwrist}, \visibility_{\leftwrist}, \visibility_{\rightwrist} \}$.
The goal is to estimate the parameters
$\params^{\text{\hmd}} = \left\{\pose \right\} \in \reals^{315}$,
that best fit $D^{\text{\hmd}}$. %
Note that we assume we are given body shape $\shape$
\cameraready{ for the \hmd fitting scenario}.

\subsection{Human Face Model and Fitting Task}
\label{subsec:human_face_model}

We represent the human face using the parametric
face model proposed by Wood et al.~\cite{Wood_2021_ICCV}.
It is a blendshape model \cite{3DMM_survey},
with $\numverts=7667$ vertices,
4 skeleton joints (head, neck and two eyes),
with their rotations and translations denoted with $\pose$, 
identity $\shape \in \reals^{256}$ and 
expression $\expression \in \reals^{233}$ blendshapes.
The deformed face mesh is obtained with standard linear
blend skinning.

\setlength\intextsep{0.0pt}
\begin{wrapfigure}[13]{l}{0.38\textwidth}
    \begin{center}
    \includegraphics[trim=065mm 018mm 065mm 015mm, clip=true,keepaspectratio,height=08em]{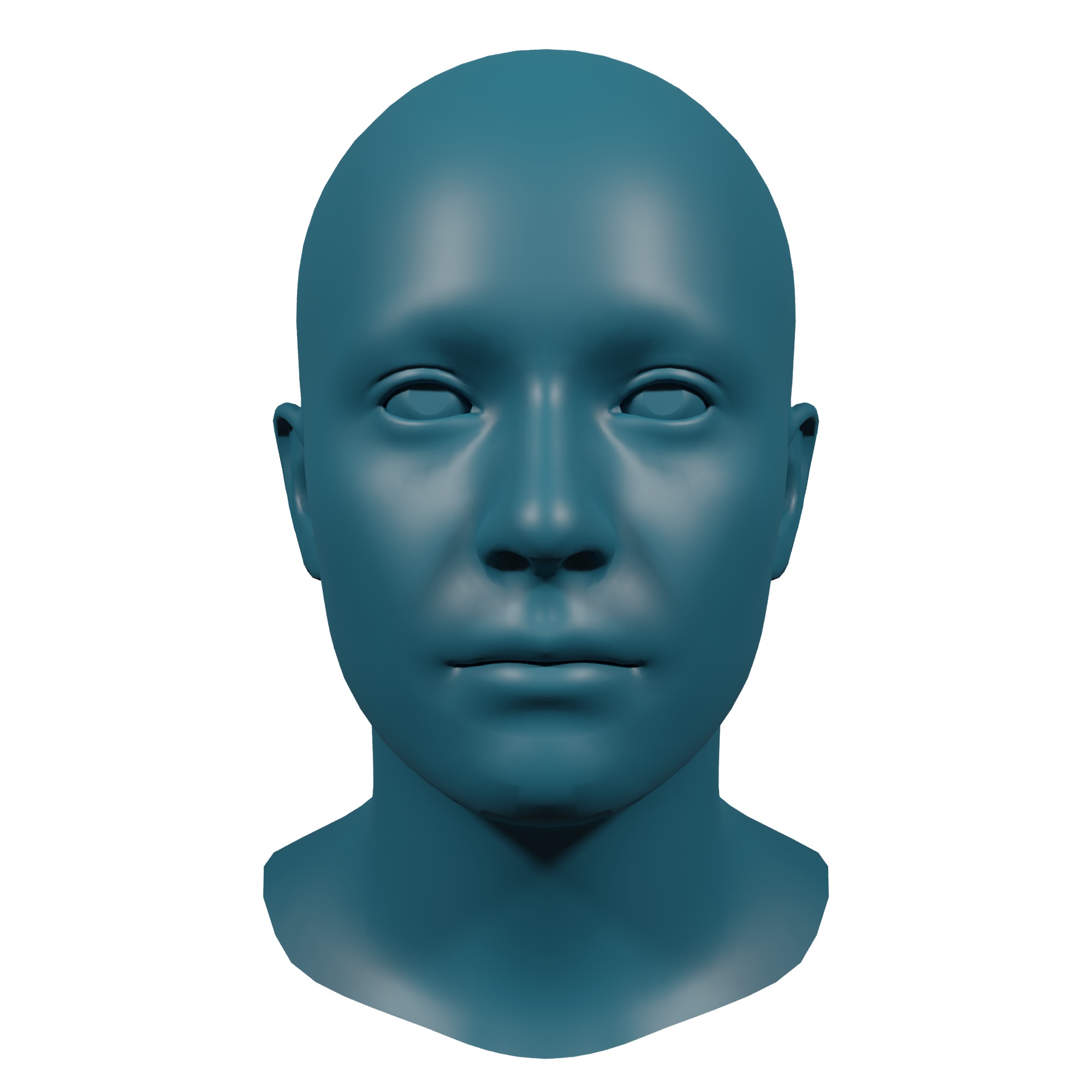}%
    \includegraphics[trim=040mm 018mm 065mm 012mm, clip=true,keepaspectratio,height=08em]{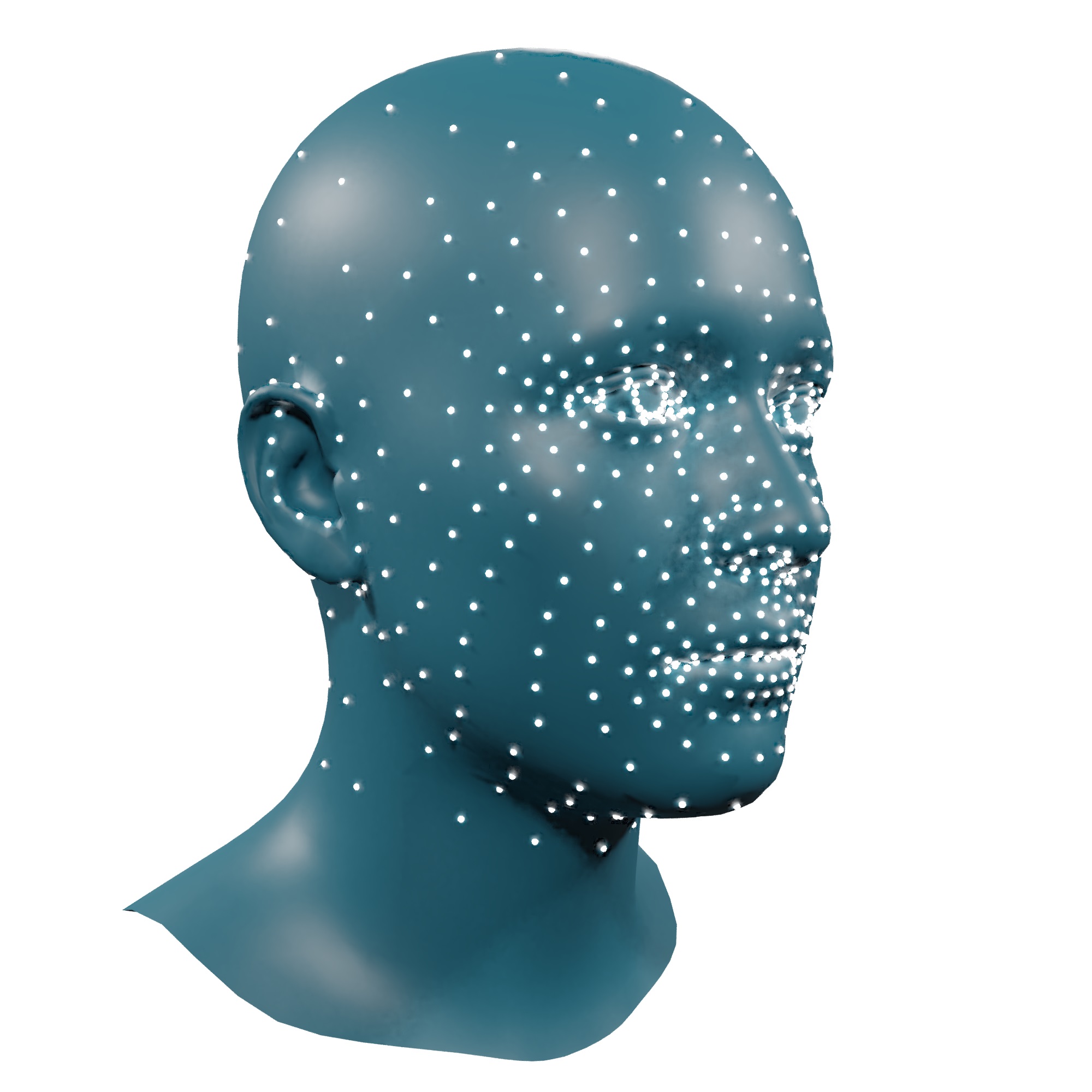}
    \end{center}
    \caption{%
    Blue: The face model template of Wood \etal \cite{Wood_2021_ICCV}.
    White: 669 dense landmarks.
    }
    \label{fig:face_model}
\end{wrapfigure}

For face fitting, we select a set of mesh vertices
as the face landmarks $\landmarks(\pose, \expression, \shape) \in \reals^{\numlandmarks \times 3}$,
$\numlandmarks = 669$ (see \fig{fig:face_model} right). 
The input data are the corresponding
\twoD face landmarks $\hat{\landmarkstwod} \in \reals^{\numlandmarks} \times 2$,
detected using the landmark neural network
proposed by Wood \etal~\cite{Wood_2021_ICCV}. %

For this task, our goal is to estimate
translation, joint rotations, expression and identity coefficients
$\params^F = \left\{\pose, \expression, \shape \right\} \in \reals^{516}$
that best fit the \twoD landmarks $D^{F} = \hat{\landmarkstwod}$.
We assume we are dealing with calibrated cameras and thus have access
to the camera intrinsics $\intrinsics$. 
$\camerapersp(\landmarks; \intrinsics)$ is 
the perspective camera projection function used to project the \threeD
landmarks $\landmarks$ onto the image plane.

\subsection{Data Terms}
\label{subsec:data_terms}

The data term is 
a function $\internalloss(\params; \data)$
that measures the discrepancy
between the observed inputs $\data$
and the parametric model evaluated at the estimated parameters $\params$. 

At the $n$-th iteration of the fitting process, we compute both 1) the array $\residuals(\params_n)$ that contains all the corresponding residuals
of the data term $\internalloss$ for the current set of parameters $\params_n$, and 2) the gradient $\grad_n = \nabla \internalloss (\params_n)$.

Let $\distance{}$ by any metric
appropriate for $\sethree$ \cite{dittadi2021full-body}
and $\rnorm{}$ a robust norm \cite{BarronCVPR2019}.
To compute residuals,
we use the Frobenius norm for $\distance{}$ and $\rnorm{}$
Note that any other norm choice can be made compatible with LM \cite{zach_eccv14}.

\qheading{Body fitting to \twoD keypoints:} We employ
the re-projection error between
the detected joints
and those estimated from the model
as the data term:
\begin{equation}
\internalloss(\params^B; D^{B}) = \rnorm{\hat{\jointstwod} - 
\camerapersp\left(\joints(\params^B), \intrinsics^B \right)
}
\label{eq:body_internal_loss}
\end{equation}
\noindent Here $\joints(\params^B)$ denotes the ``posed'' joints.

\qheading{Body fitting to \hmd signals:} We measure the discrepancy
between the observed data $\data^{\text{\hmd}}$ and the estimated model parameters
$\params^{\text{\hmd}}$ with the following data term:
\begin{equation}
\begin{aligned}
&\internalloss(\params^{\text{\hmd}}; \data^{\text{\hmd}}) = 
\distance{\hat{T}^{\head}, T^{\head}(\params^{\text{\hmd}})} + \\
 & \sum_{w \in {\leftwrist,\rightwrist} }v_w \left( \distance{\hat{T}^w, T^w(\params^{\text{\hmd}})} +
 \sum_{i=1}^5 \rnorm{\hat{P}_i^w - P_i^w(\params^{\text{\hmd}})}
 \right)
\end{aligned}\label{eq:internal_loss}
\end{equation}

\qheading{Face fitting to \twoD landmarks:}
\cameraready{The data term is the landmark re-projection error:}
\begin{equation}
\internalloss(\params^F; D^{F}) = \rnorm{\hat{\landmarkstwod} - 
\camerapersp\left(\landmarks(\params^F) ; \intrinsics^F \right)
}
\label{eq:face_internal_loss}
\end{equation}

\subsection{Training Details}
\label{subsec:training_details fitter}

\qheading{Training losses:}
\cameraready{We train our learned fitter using a combination of model parameter and mesh
losses. Their precise formulation can be found in the \supmat}

\qheading{Model structure:}
Unless otherwise specified,
$f, f_{\damping,\lr}$
(in \cref{alg:fitting},
\eqref{eq:gamma_damping}) use a stack of two \gru{s}
with 1024 units each. The initialization $\initregressor, \Phi_h$ in \cref{alg:fitting} are MLPs with two
layers of 256 units, \relu
\cite{nair2010rectified} and Batch Normalization
\cite{ioffe2015batch}.

\qheading{Datasets:} 
For the body fitting tasks, we use \amass \cite{AMASS:ICCV:2019}
to train and test our fitters.
When fitting \smpl to \twoD keypoints,
we use \threedpw's \cite{vonMarcard2018} test set
to evaluate the learned fitter's accuracy,
\cameraready{using the detected OpenPose \cite{openpose} keypoints as the target}.
The face fitter is trained and evaluated on synthetic data%
.
Please see the \supmat for more details on the datasets.

\section{Experiments}
\label{sec:experiments}

\subsection{Metrics}
\label{subsec:metrics}

Metrics with a \emph{PA} prefix are computed after undoing rotation, scale
and translation, \ie Procrustes alignment.
Variables with a \emph{tilde} are {ground-truth} values.

\qheading{Vertex-to-Vertex (\vtov):}
As we know the correspondence
between ground-truth $\gt{\mesh}$
and estimated vertices $\mesh$,
we are able to compute the mean per-vertex error: $    \vtov(\gt{\mesh}, \mesh) = \frac{1}{\numverts} 
    \sum_{i=1}^{\numverts} \normmse{\gt{\mesh}_i - \mesh_i
    }$.
 For \smplh, in addition to the full mesh error (FB), we report
error values
for the head (H) and hands (L, R).
A visualization of the selected parts
is included in the \supmat
The \textbf{3D per-joint error (\mpjpe)} is equal to:
$\mpjpe(\gt{\joints}, \joints) = \frac{1}{\numjoints} 
    \sum_{i=1}^{\numjoints} \normmse{\gt{\joints}_i - \joints_i
    }$.

\qheading{Ground penetration (\groundmetric):} We report
the average distance to the ground plane for all vertices
below ground \cite{yuan2021simpoe}:
$\text{\groundmetric}(\mesh) = \frac{1}{\lvert \mathtt{S} \rvert} \sum_{%
    n \in \mathtt{S} 
    } \lvert d_{\ground}(M_i) \rvert$, where $d_{\ground}({\mesh}_i) = \mesh_i \cdot n_{\ground}$ and $\mathtt{S} = \{i \mid d_{\ground}(\mesh_i) < 0 \}$.

\label{subsec:datasets}

\qheading{Face landmark error (\mplpe):}
We report the mean distance
between estimated and ground-truth \threeD landmarks
$\mplpe(\gt{\landmarks}, \landmarks) = \frac{1}{\numlandmarks} 
    \sum_{i=1}^{\numlandmarks} \normmse{\gt{\landmarks}_i - \landmarks_i
}$.

\begin{table}[t]
    \captionsetup{font=footnotesize}
    \caption{%
Using \threedpw \cite{vonMarcard2018} to compare different approaches that estimate \smpl
from images, \twoD keypoints and part segmentation masks.
Replacing LGD's \cite{jie_eccv_2020} update rule with ours
leads to a 2 mm PA-MPJPE improvement.
Our full system, that uses \gru{s}, leads to a further 1.6 mm improvement.
 “O/R” denotes Optimization/Regression.
    }
    \renewcommand{\arraystretch}{\myarraystretch}
\begin{center}
    \begin{tabular}{l|ccccc}
    Method & Type & Image & \twoD keypoints & Part segmentation & \pampjpe \\ 
    \toprule 
    \smplify \cite{Bogo:ECCV:2016} & O & \xmark & \cmark & \xmark & 106.1 \\
    SCOPE \cite{Fan_2021_ICCV} & O & \xmark & \cmark & \xmark & 68.0 \\
    SPIN \cite{Kolotouros_2019_ICCV} & R & \cmark & \xmark & \xmark & 59.6 \\
    VIBE \cite{VIBE:CVPR:2020} & R & \cmark & \xmark & \xmark & 55.9 \\
    Neural Descent \cite{Zanfir_2021_CVPR} & R+O & \cmark & \cmark & \cmark & 57.5 \\ 
    LGD \cite{jie_eccv_2020} & R+O & \xmark & \cmark & \xmark & 55.9 \\ 
    Ours, LGD + \eq{eq:line_search_like} & R+O & \xmark & \cmark & \xmark & 53.9 \\
    Ours (full) & R+O & \xmark & \cmark & \xmark & \textbf{52.2} \\ 
    \bottomrule
    \end{tabular}
    \end{center}
    
    \label{tab:body_twod}
\end{table}

\subsection{Quantitative Evaluation}
\label{subsec:quant}

\begin{table}[b]

    \captionsetup{font=footnotesize}
    \caption{Fitting \smplh to simulated sequences of
        \hmd data. %
        Our proposed fitter outperforms the classical optimization baselines
        (\optimname prefix)
        on the full body and ground penetration metrics, with similar or better performance
        on the part metrics,
        and the regressor baselines (the
        VAE predictor \cite{dittadi2021full-body} and
        the regressor $\initregressor$), on all metrics.
        “F/H” denotes full / half-plane visibility.
    }
    \renewcommand{\arraystretch}{\myarraystretch}
    \scriptsize
    \centering
    \begin{tabular}{l|cc|cc|cc|cc|cc}
          & \multicolumn{6}{c|}{\vtovlong (mm) $\downarrow$}
          & \multicolumn{2}{c|}{\mpjpe}
          & \multicolumn{2}{c}{\groundmetric}
        \\
        Method
          & \multicolumn{2}{c|}{Full body}
          & \multicolumn{2}{c|}{Head}
          & \multicolumn{2}{c|}{L / R hand }
          & \multicolumn{2}{c|}{(mm) $\downarrow$}
          & \multicolumn{2}{c}{(mm) $\downarrow$}
        \\
          &                                                                                  %
        F & H                                                                                %
          & F                                                & H                             %
          & F                                                & H                             %
          & F                                                & H                             %
          & F                                                & H                             \\ %
        \toprule
        \optimname, GMM
          & 73.1                                             & 116.2                         %
          & 2.9                                              & 3.4                           %
          & 3.2 / 3.0                                        & 5.6 / 5.3                     %
          & 49.7                                             & 137.26                        %
          & 70.8                                             & 74.0                          \\
        \optimname, GMM, Tempo.
          & 72.6                                             & 113.3
          & 2.9                                              & 3.4
          & 3.3 / 3.1                                        & 6.8 / 6.5
          & 49.4                                             & 132.1
          & 70.7                                             & 73.5                          \\
        \optimname, VAE Enc.
          & 76.1                                             & 119.3
          & 3.9                                              & 4.1
          & 5.3  / 4.7                                       & 8.7 / 7.6
          & 52.6                                             & 140.5
          & 63.6                                             & 66.7                          \\
        Dittadi \etal \cite{dittadi2021full-body}
          & \multicolumn{2}{c|}{\na}
          & \multicolumn{2}{c|}{\na}
          & \multicolumn{2}{c|}{\na}
          & 43.3                                             & \na
          & \multicolumn{2}{c}{\na}
        \\
        Ours $\initregressor, (N = 0)$
          & 44.2                                             & 69.7
          & 19.1                                             & 22.7
          & 27.8 / 25.9                                      & 32.1 / 29.9
          & 38.9                                             & 84.9
          & 16.1                                             & 20.1                          \\
        Ours $(N = 5)$
          & \textbf{26.1}                                    & \textbf{49.9}
          & \textbf{2.2}                                     & \textbf{3.2}
          & \textbf{3.0} / \textbf{3.3}                      & \textbf{3.1}  /  \textbf{3.7}
          & \textbf{18.1}                                    & \textbf{62.1}
          & \textbf{12.5}                                    & \textbf{15.5}                 \\
        \bottomrule
    \end{tabular}
    \label{tab:main}
\end{table}

\qheading{Fitting the body to \twoD keypoints:}
We compare our proposed update
rule with existing regressors, classic and learned optimization methods
on \threedpw~\cite{vonMarcard2018}. For a fairer comparison with Song \etal \cite{jie_eccv_2020},
we train two versions of our proposed fitter, one where we change the update
rule of \lgd
with \eq{eq:line_search_like},
and our full system which also has network architecture changes.
\Tab{tab:body_twod} shows that just by changing the update rule
(Ours, \lgd + \eq{eq:line_search_like}),
we outperform all baselines. %
\\
\qheading{Fitting the body to \hmd data:}
In \tab{tab:main} we compare our proposed learned optimizer with a standard optimization
pipeline,
a variant of \smplify \cite{Bogo:ECCV:2016,Pavlakos2019_smplifyx} adapted to the \hmd fitting task
(first 3 rows), and two neural network regressors
(a VAE predictor \cite{dittadi2021full-body} in the 4th row
and our initializer $\initregressor$ of \alg{alg:fitting} in the 5th row),
on the task of fitting \smplh to sparse \hmd signals, see \cref{subsec:human_model}%
.
The optimization baseline minimizes the energy with data term ($\internalloss$ in \cref{eq:internal_loss}), gravity term $\gravityloss$, prior term $\poseprior$, without/with temporal term $\loss^{\temporal}$ (first/second row of \tab{tab:main})
to estimate the parameters $\params_{1, \ldots, \temporal }$ of a sequence
of length $\temporal$:
\begin{equation}
\begin{aligned}
    &\loss^{\optim}(\params^{\text{\hmd}}) = \internalloss(\params^{\text{\hmd}}; \data^{\text{\hmd}}) + \gravityloss + \poseprior +
    \loss^{\temporal}
    \\
    &\gravityloss(\params^{\text{\hmd}}) =  1 - \frac{%
     T_{\text{pelvis}}(1, :3) \cdot \up
    }{%
    \normmse{T_{\text{pelvis}}(1, :3)}
    \normmse{{\up}}
    } , \, \up = (0, 1, 0) \\
    &  \loss^{\temporal} (\params^{\text{\hmd}})= \sum_{t=1}^{\temporal-1}\distance{
    \transform_{t+1}(\params_{t+1}^{\text{\hmd}}) - \transform_{t}(\params_{t}^{\text{\hmd}})
    }
\end{aligned}
\end{equation}
We use two different pose priors, a GMM \cite{Bogo:ECCV:2016} and
a VAE encoder $\encoder(*)$ \cite{Pavlakos2019_smplifyx}:
\begin{align}
    & \posepriorgmm = - \min_{j}{%
    \log{%
    \left( w_j \mathcal{N}(\pose; \mu_{\pose, j}, \Sigma_{\pose, j}) \right)}
    }
    \\
    & \posepriorvae = \text{Neg. Log-Likelihood}(\mathcal{N}(\encoder(\pose), \mathcal{I}) )
\end{align}

We minimize the loss above using L-BFGS \cite[Ch.~7.2]{NoceWrig06}
for 120 iterations on the test split of the \mocap data.
We choose L-BFGS instead of Levenberg-Marquardt, since PyTorch currently
lacks the feature to efficiently compute jacobians, without having to resort
to multiple backward passes for derivative computations.
We report
the results for both full and half-space visibility
in \tab{tab:main} using the metrics of \Sect{subsec:metrics}.
Our method outperforms the baselines in terms of full-body and penetration metrics,
and shows competitive performance \wrt to the part metrics.
Regression-only methods \cite{dittadi2021full-body}
cannot tightly fit the data,
due to the lack of a feedback mechanism.

\begin{figure}[t]
    \begin{center}
        \includegraphics[trim=000mm 000mm 000mm 000mm, clip=true, width=0.33\linewidth]{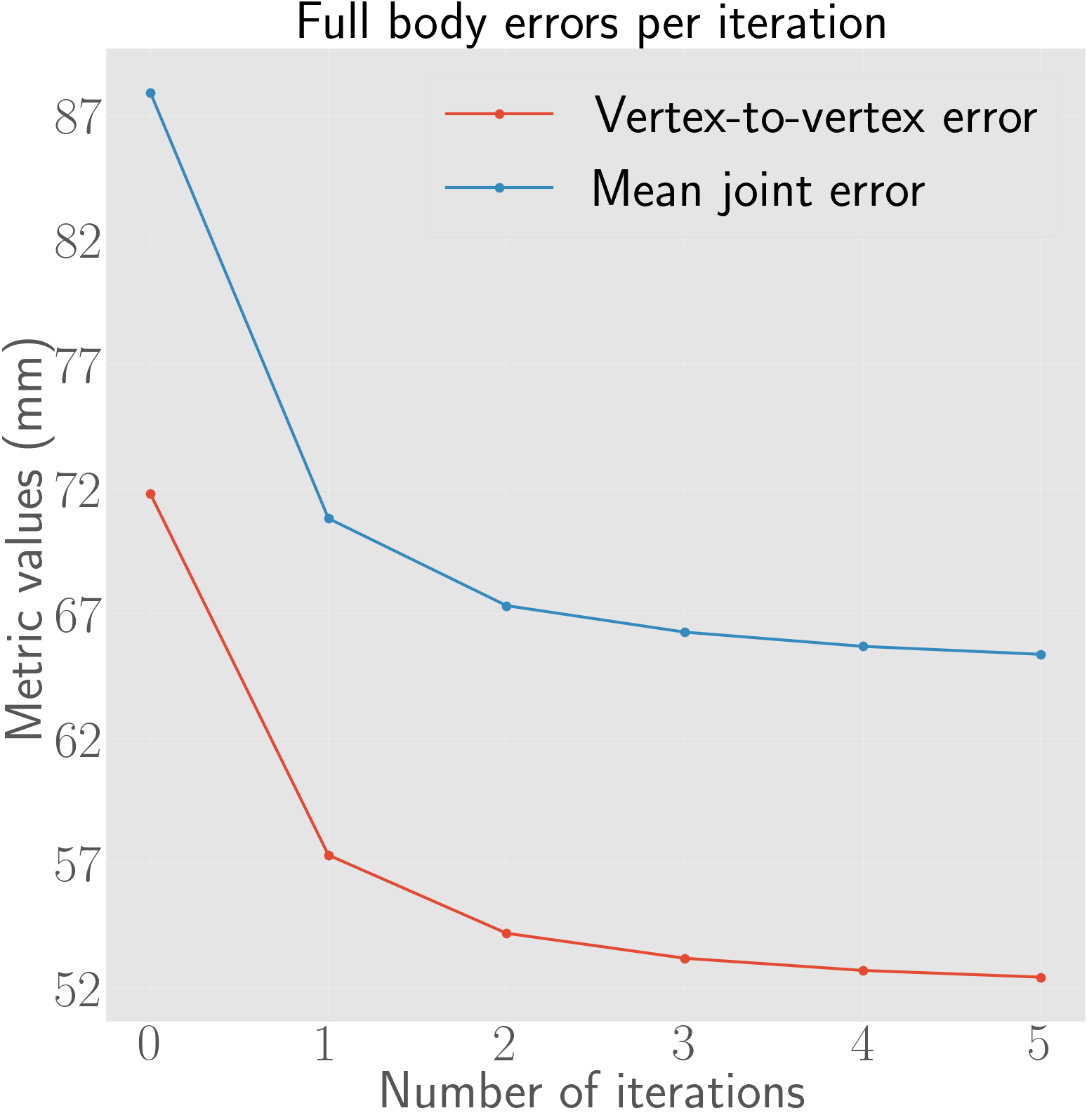}%
        \includegraphics[trim=000mm 000mm 000mm 000mm, clip=true, width=0.33\linewidth]{images/half_plane_error_per_iter/parts.pdf}%
        \includegraphics[trim=000mm 000mm 000mm 000mm, clip=true, width=0.33\linewidth]{images/half_plane_error_per_iter/gnd_penetration.pdf}
    \end{center}
    \caption{%
        Errors per iteration when fitting \smplh
        to \hmd data for the half-space visibility scenario,
        see \supmat for full visibility.
        Left to right: 1) full body vertex and joint errors, 2) head, left and
        right hand \vtov errors and 3) vertex and joint ground distance,
        computed on the set of points below ground.
    }

    \label{fig:error_per_iter}
\end{figure}

\qheading{Runtime:} Our method (PyTorch) runs at \emph{150 ms} per frame
on a P100 GPU, while the baseline \optimname
method (PyTorch) above requires \emph{520 ms}, on the same hardware. We are aware that a highly optimized real-time version of the latter exists and runs at 
\emph{0.8 ms} per frame, performing at most 3 \lm iterations, but
it requires investing significant effort into a problem specific C++ codebase.%

\fig{fig:error_per_iter} contains the metrics per iteration of our method,
averaged across the entire test dataset.
It shows that our learned fitter is able to 
aggressively optimize the target data term and converge quickly.

\ifdef{\captionsharedweights}{}{
\newcommand{\captionsharedweights}[0]{%
    Using per-step network weights reduces
    head and ground penetration
    errors,
    albeit at an N-folder parameter increase.
}
}
\ifdef{\networkchoice}{}{
\newcommand{\networkchoice}[0]{%
   \gru vs a residual feed-forward network
    \cite{he_resnets,highway}.
    \gru{'s} memory makes it more effective. Multiple layers bring further benefits,
    but increase runtime.
}
}

\begin{table}[t]
  \captionsetup{font=footnotesize}
  \mbox{}\hfill
  \begin{minipage}[t]{.48\linewidth}
  \caption{\captionsharedweights}
  \renewcommand{\arraystretch}{\myarraystretch}
  \resizebox{1.00\linewidth}{!}{
  \centering
  \begin{tabular}{l|ccc|c|c}
Weights
& \multicolumn{3}{c|}{\vtov (mm) $\downarrow$}
& \multicolumn{1}{c|}{\mpjpe}
& \multicolumn{1}{c}{\groundmetric} 
\\
    & \fullbody & \headshort & \leftrighthands & 
    (mm) $\downarrow$ & (mm)$\downarrow$
   \\
    \toprule
    Shared  & 52.3 & 3.5 & 3.6 / \textbf{3.7} & 64.1 & 18.2 \\
    Per-step  &  \textbf{49.9} & \textbf{3.2}  & \textbf{3.1}  /  \textbf{3.7} & \textbf{62.1} &
    \textbf{15.5} \\
    \bottomrule
\end{tabular}
  }
  \label{tab:shared_weights}
  \end{minipage}\hfill
  \begin{minipage}[t]{.48\linewidth}
  \caption{%
    \networkchoice
    }
    \renewcommand{\arraystretch}{\myarraystretch}
    \resizebox{1.00\linewidth}{!}{
    \centering
    \begin{tabular}{l|ccc|c|c}
    Network
    & \multicolumn{3}{c|}{\vtov (mm) $\downarrow$}
    & \multicolumn{1}{c|}{\mpjpe}
    & \multicolumn{1}{c}{\groundmetric} 
    \\
        Structure & \fullbody & \headshort & \leftrighthands & 
        (mm) $\downarrow$ & (mm)$\downarrow$
       \\
        \toprule
        \resnet  & 65.3 & 6.8 & 7.3 / 7.6 & 73.1 & 16.2 \\
        \gru(1024)  & 53.6 & 3.7 & 3.4 / 4.0 & 66.1 & \textbf{15.1} \\
        \gru(1024, 1024) & \textbf{49.9} & \textbf{3.2}  & \textbf{3.1}  /  \textbf{3.7} & \textbf{62.1} &
        15.5 \\
        \bottomrule
    \end{tabular}
    
    }
    
    \label{tab:network_structure}
  \end{minipage}\hfill
  \mbox{}
\end{table}
\ifdef{\captionupdaterule}{}{
\newcommand{\captionupdaterule}[0]{%
    Comparison of our update rule (\eq{eq:line_search_like}) with the pure network update $\Delta\Theta_{n}$.
    Our proposed combination 
    improves the results for all metrics.
    
}
}
\ifdef{\captiongdlr}{}{
\newcommand{\captiongdlr}[0]{%
Learning to predict $\gamma$ is better than a constant, with 
performance degrading gracefully, providing an option for a lower
computational cost.
}
}

\begin{table}[t]
  \captionsetup{font=footnotesize}
  \mbox{}\hfill
  \begin{minipage}[t]{.48\linewidth}
  \caption{\captionupdaterule}
\renewcommand{\arraystretch}{\myarraystretch}
    \resizebox{1.00\linewidth}{!}{
    \centering

    \begin{tabular}{l|ccc|c|c}
    Update
    & \multicolumn{3}{c|}{\vtov (mm) $\downarrow$}
    & \multicolumn{1}{c|}{\mpjpe}
    & \multicolumn{1}{c}{\groundmetric} 
    \\
    Rule & \fullbody & \headshort & \leftrighthands & 
    (mm) $\downarrow$ & (mm)$\downarrow$
    \\
    \toprule
    +$\Delta\Theta_{n}$ & 53.8 & 14.7 & 7.8 / 7.9 & 66.3 & 15.8 \\
    +\eq{eq:line_search_like}
    & \textbf{49.9} & \textbf{3.2}  & \textbf{3.1}  /  \textbf{3.7} & \textbf{62.1} & \textbf{15.5}  \\
    \bottomrule
    \end{tabular} 
    }
    
    \label{tab:update_rule}
  
  \end{minipage}\hfill
  \begin{minipage}[t]{.48\linewidth}
  \caption{\captiongdlr}
    \renewcommand{\arraystretch}{\myarraystretch}
    \resizebox{1.00\linewidth}{!}{
    \centering
    \begin{tabular}{l|lll|l|l}
    Learning
    & \multicolumn{3}{c|}{\vtov (mm) $\downarrow$}
    & \multicolumn{1}{c|}{\mpjpe}
    & \multicolumn{1}{c}{\groundmetric} 
    \\
        rate $\lr$ & \fullbody & \headshort & \leftrighthands & 
        (mm) $\downarrow$ & (mm)$\downarrow$
       \\
        \toprule
        \text{1e-4} & 51.9 & 3.5 & 3.8 / 4.6 & 64.2 & \textbf{15.5} \\
        Learned & \textbf{49.9} & \textbf{3.2}  & \textbf{3.1}  /  \textbf{3.7} & \textbf{62.1} & \textbf{15.5}  \\
        \bottomrule
\end{tabular}
    }
    
    \label{tab:learning_rate}
  \end{minipage}\hfill
  \mbox{}
\end{table}

\qheading{Ablation study:} We perform our ablations
on the problem
of fitting \smplh to \hmd signals, using the  
half-space visibility setting. %
Unless otherwise stated, we report the performance
\cameraready{
of regression and 5 iterations of the learned fitter.
}

We first compare two variants of the fitter, one with shared and the other with 
separate network weights per optimization step.
\Tab{tab:shared_weights} shows that the latter can help reduce the errors,
at the cost of an N-fold increase in memory.

Secondly, we investigate the effect of the type and structure of 
the network, replacing the \gru with
a feed-forward network with skip
connections, i.e., ResNet \cite{he_resnets,highway}. 
We also train a version of our fitter with a single \gru
with 1024 units. %
\Tab{tab:network_structure} shows that the \gru is better suited to
this type of problem, thanks to its internal memory.
This is very much in line with many popular continuous optimizer
work \cite{Zanfir_2021_CVPR}.

Thirdly, we compare the update rule
of \eq{eq:line_search_like} with 
a learned fitter
that only uses
the network update, \ie 
$\lr=0, \damping=1$ in \eq{eq:line_search_like}.
This is an instantiation of \lgd \cite{jie_eccv_2020},
albeit with a different network and task.
\Tab{tab:update_rule} shows that the proposed weighted combination is better
than the pure network update.

Fourthly, we investigate whether we need to learn the step
size $\lr$ or if a constant value is enough.
\Tab{tab:learning_rate} shows that performance gracefully degrades
when using a constant learning value. Therefore, it is an
option for decreasing the computational cost, without a significant performance drop.

Finally, we present some qualitative results in \fig{fig:hmd_qualitative}. Notice
how the learned fitter corrects the head pose and hand articulation of the initial 
predictions.

\ifdef{\captionfaces}{}{
\newcommand{\captionfaces}[0]{%
    Face fitting to 
    \twoD landmarks.
}
}

\setlength{\intextsep}{0pt}
\begin{wraptable}[10]{r}{0.4\textwidth}
    \captionsetup{font=footnotesize}
    \vskip -0mm
    \caption{\captionfaces}
    \renewcommand{\arraystretch}{\myarraystretch}
    \resizebox{1.00\linewidth}{!}{
    \begin{tabular}{l|cc cc|cc}
& \multicolumn{4}{c|}{\vtov (mm) $\downarrow$}
& \multicolumn{2}{c}{\mplpe}
\\
&  \multicolumn{2}{c|}{Face} & \multicolumn{2}{c|}{Head} & \multicolumn{2}{c}{ (mm) $\downarrow$} \\
Method 
& - & PA  
& - & PA
& - & PA  \\
\toprule 
LM
& 34.4 & 3.7 
& 33.8 & 5.3
& 33.8 & \textbf{3.4}
\\
Ours
& \textbf{7.9} & \textbf{3.5}
& \textbf{8.5} & \textbf{4.1}
& \textbf{8.0} & 3.7
\\
\bottomrule
\end{tabular}
    }
    \label{tab:faces}
\end{wraptable}
\qheading{Face fitting to \twoD landmarks:}
We compare our proposed learned optimizer with a C++ production grade solution that uses \lm to solve the face fitting problem
described in Sec.~\ref{subsec:human_face_model}. Given the per-image 2D landmarks as input, the optimization baseline minimizes the energy with data term ($\internalloss$ in \eq{eq:face_internal_loss}) and a simple regularization term to estimate $\params^F = \left\{\pose, \expression, \shape \right\}$:
\begin{align}
    &\loss^{\optim}(\params^{F}) = \internalloss(\params^{F}; \data^{F}) + \mathbf{w} * \normmse{\params^{F}}
\end{align}
$\mathbf{w}$ contains the different regularization weights for $\pose, \expression, \shape$, which are tuned manually for the best baseline result.

\ifdef{\captionfacequal}{}{
\newcommand{\captionfacequal}[0]{%
    Face model \cite{Wood_2021_ICCV} fitting to to dense \twoD landmarks
    a) target \twoD landmarks, b) LM fitter,
    c) ours, d) ground-truth.  
}
}

\begin{wrapfigure}[17]{l}{0.38\textwidth}
    \begin{center}
        \includegraphics[trim=080mm 050mm 050mm 070mm, clip=true,width=0.24\linewidth]{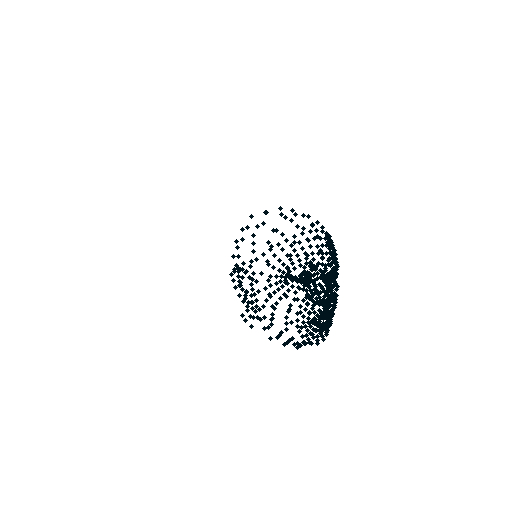}%
        \includegraphics[trim=080mm 050mm 050mm 070mm, clip=true,width=0.24\linewidth]{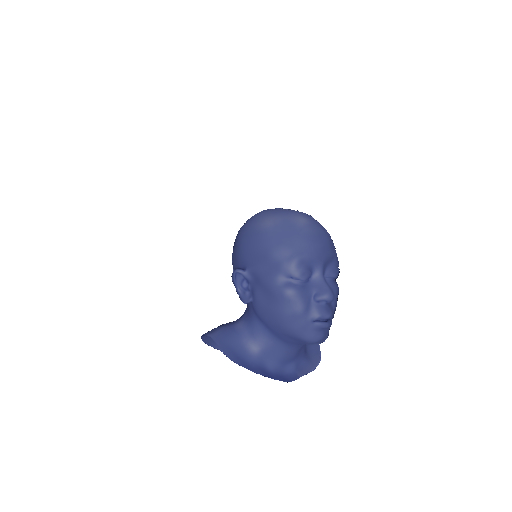}%
        \includegraphics[trim=080mm 050mm 050mm 070mm, clip=true,width=0.24\linewidth]{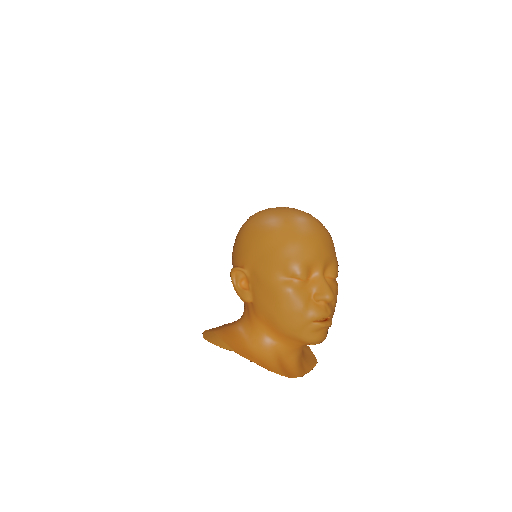}%
        \includegraphics[trim=080mm 050mm 050mm 070mm, clip=true,width=0.24\linewidth]{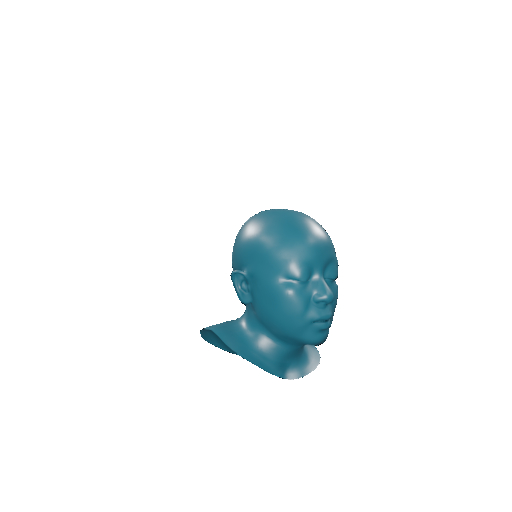}
        \includegraphics[trim=080mm 050mm 050mm 070mm, clip=true,width=0.24\linewidth]{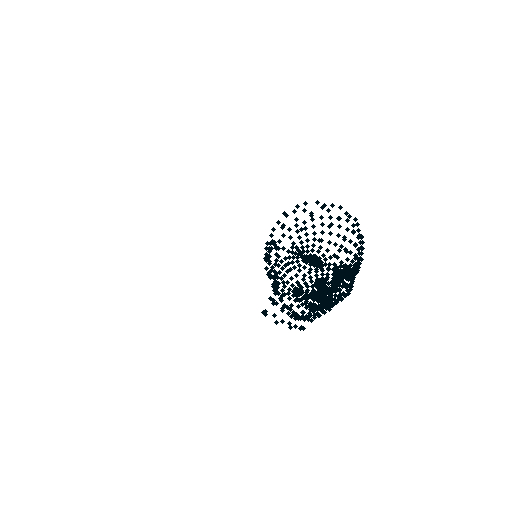}%
        \includegraphics[trim=080mm 050mm 050mm 070mm, clip=true,width=0.24\linewidth]{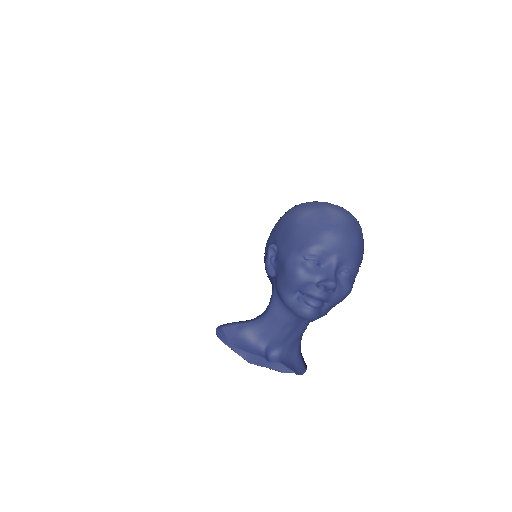}%
        \includegraphics[trim=080mm 050mm 050mm 070mm, clip=true,width=0.24\linewidth]{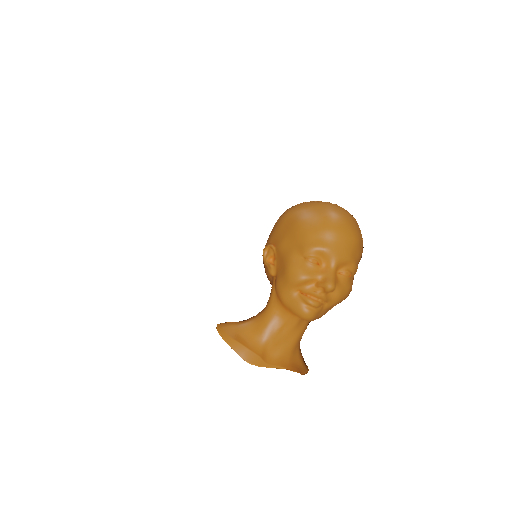}%
        \includegraphics[trim=080mm 050mm 050mm 070mm, clip=true,width=0.24\linewidth]{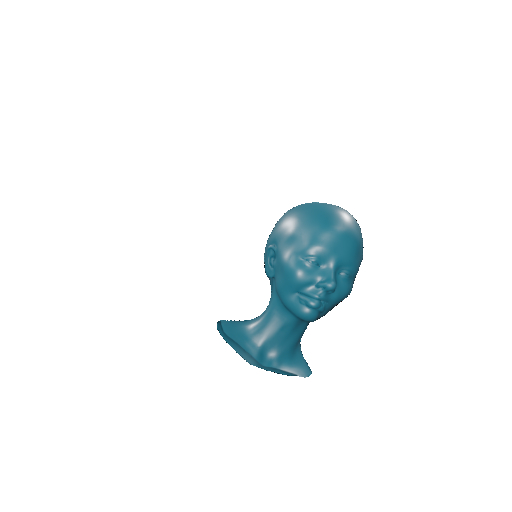}
        \includegraphics[trim=065mm 050mm 055mm 060mm, clip=true,width=0.24\linewidth]{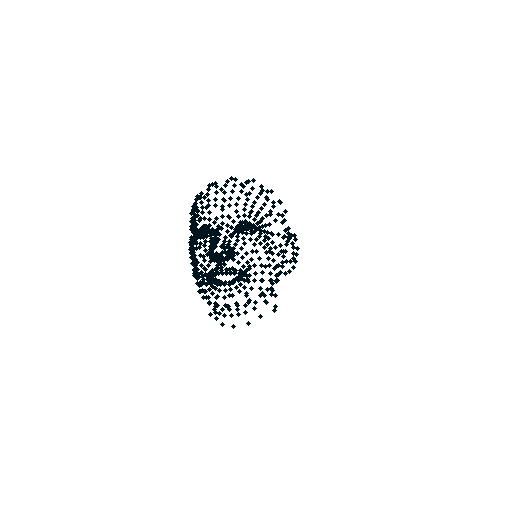}%
        \includegraphics[trim=065mm 050mm 055mm 060mm, clip=true,width=0.24\linewidth]{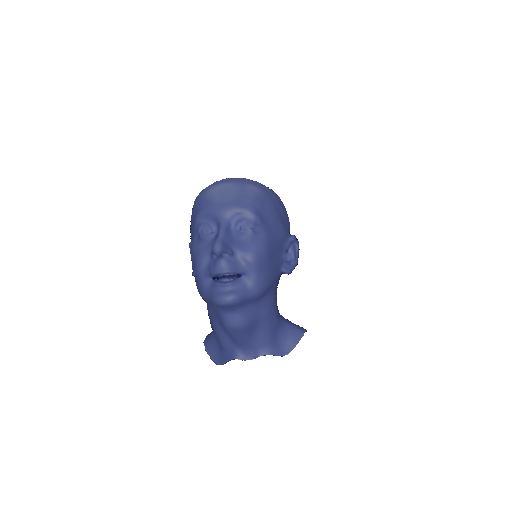}%
        \includegraphics[trim=065mm 050mm 055mm 060mm, clip=true,width=0.24\linewidth]{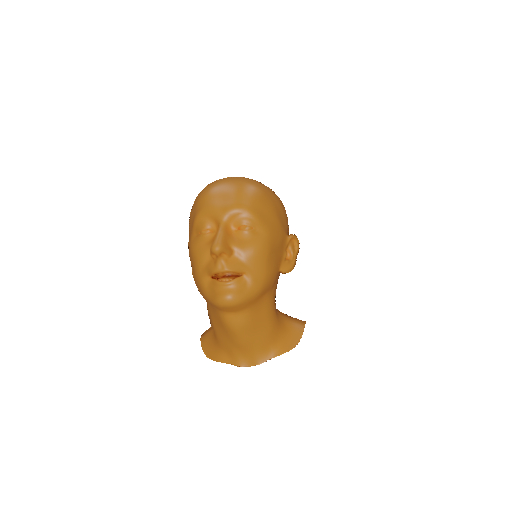}%
        \includegraphics[trim=065mm 050mm 055mm 060mm, clip=true,width=0.24\linewidth]{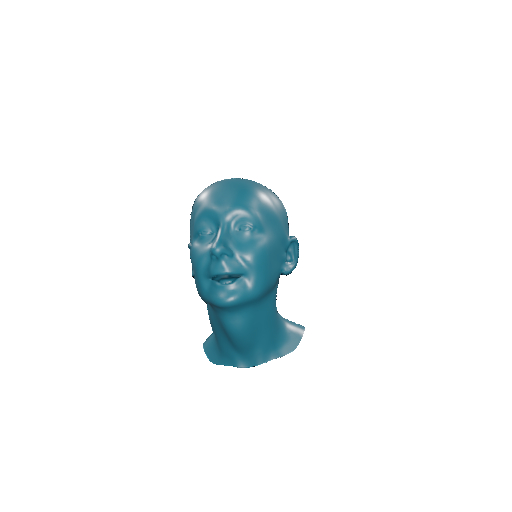}\\
        {%
        \small
        \makebox[0.24\linewidth]{a)}%
        \makebox[0.24\linewidth]{b)}%
        \makebox[0.24\linewidth]{c)}%
        \makebox[0.24\linewidth]{d)}%
        }
    \end{center}
    \setlength{\abovecaptionskip}{-0.5em}
    \caption{\captionfacequal}
    \label{fig:face_qual}
\end{wrapfigure}
The quantitative comparison in \tab{tab:faces} shows that our proposed
fitter outperforms the \lm baseline on almost all metrics.
The large value in absolute errors (``-'' columns) is due
to the wrong estimation of the depth of the mesh.
After alignment (\emph{PA} columns), the gap is much smaller.
See \fig{fig:face_qual} for a qualitative comparison.

\qheading{Runtime:} %
Here, the baseline optimization is in C++ and thus for a fair comparison, we only compare the time it takes to compute the parameter update given the residuals and jacobians (per-iteration).
Computing the values of the learned parameter update (ours, using PyTorch)
takes \emph{12 ms} on a P100 GPU, while
computing
the \lm update (baseline, C++) requires
\emph{34.7 ms} ($504$ free variables).
Note that the \lm update
only
requires \emph{0.8 ms} on a laptop CPU when optimizing over $100$ free variables.
The difference is due to the cubic complexity of \lm \wrt
the number of free variables of the problem.

\ifdef{\captionhmdqual}{}{
\newcommand{\captionhmdqual}[0]{%
    Estimates in yellow, ground-truth in blue, best viewed in color.
     Our learned optimizer successfully fits the target
     data
     and produces plausible poses for the full \threeD body.
     Points that are greyed out are outside of the field of view,
     \eg the hands in the second row, and thus not perfectly fitted.
}
}

\begin{figure}[t]
  \begin{center}
  \includegraphics[width=0.99\linewidth]{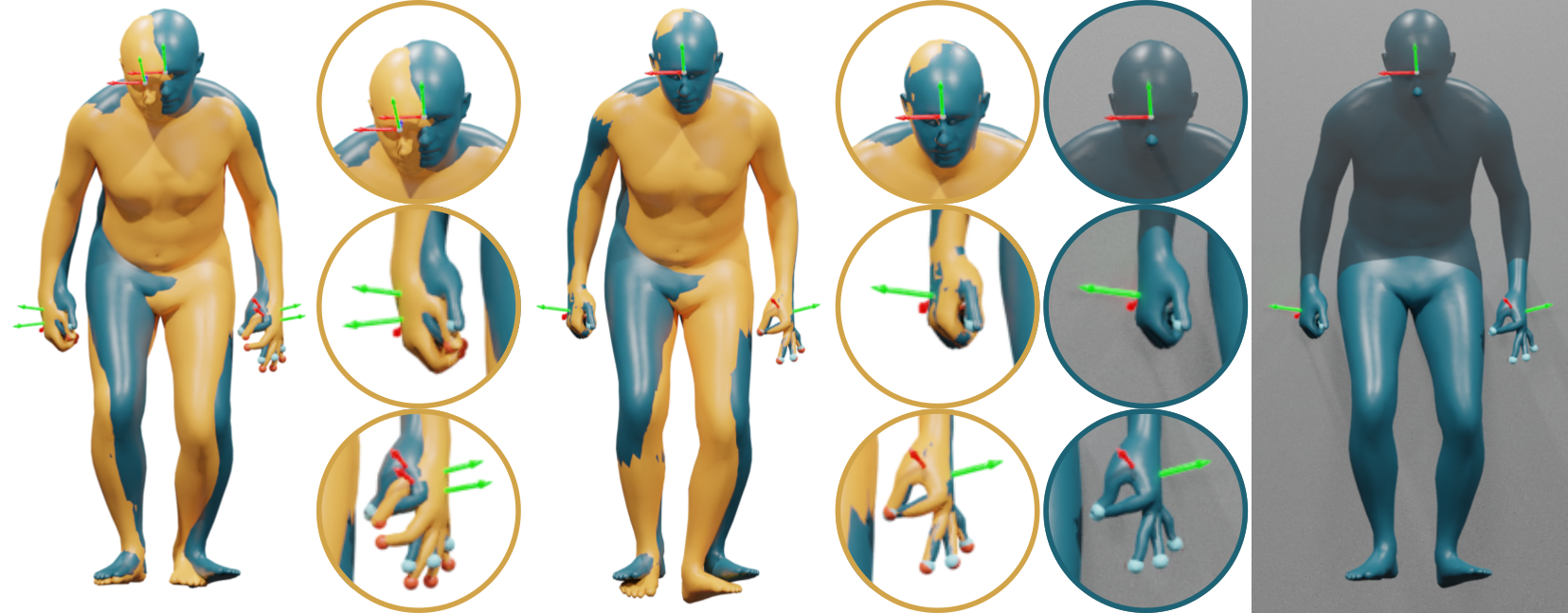}
  \includegraphics[width=0.99\linewidth]{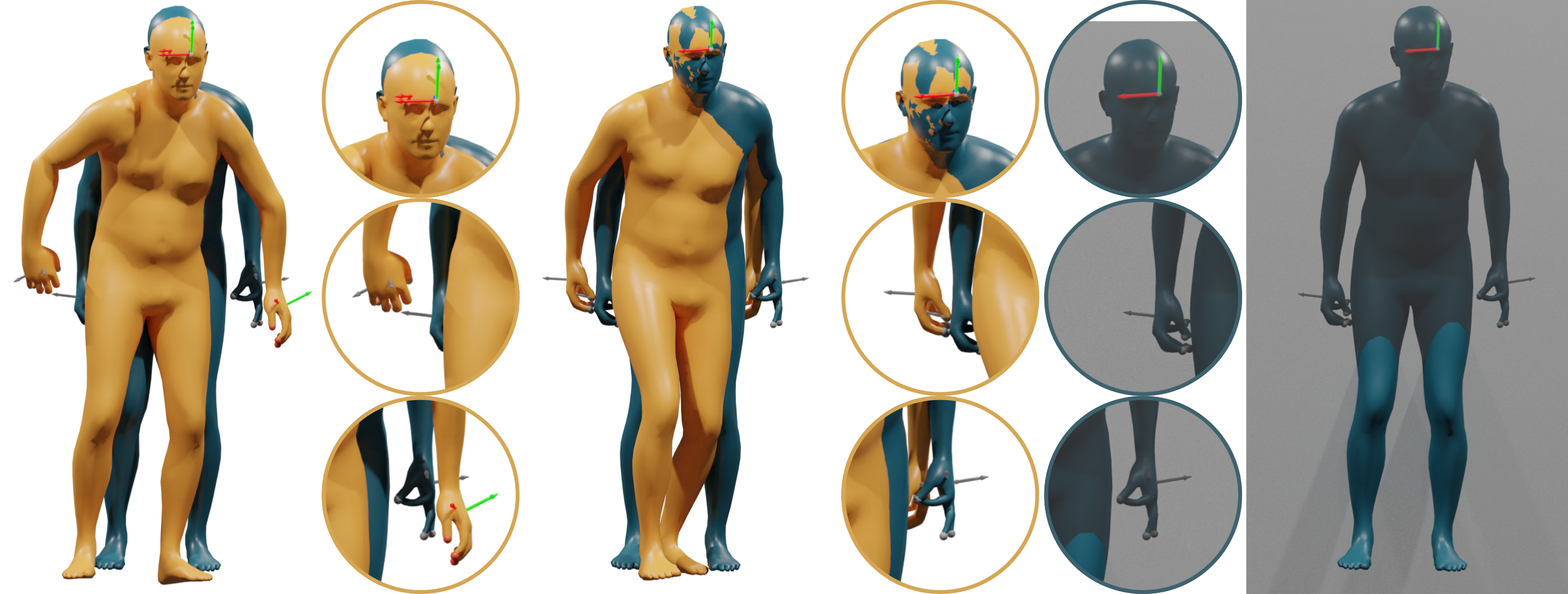}
  \end{center}
    {%
    \normalsize
    \makebox[0.33\linewidth]{1) Initial $\initregressor$ output}%
    \makebox[0.33\linewidth]{2) Iteration $N = 5$}%
    \makebox[0.33\linewidth]{3) Ground-truth}%
    }
   \caption{\captionhmdqual}
    \label{fig:hmd_qualitative}
\end{figure}
\subsection{Discussion}
\label{subsec:discussion}

If we apply the proposed method to a sequence of data, we will
get plausible per-frame results, but the overall motion will be implausible.
Since the model is trained on a per-frame basis and lacks temporal context,
it cannot learn the proper dynamics present in temporal data.
Thus, limbs in successive frames will move unnaturally, with large jumps
or jitter. Future extensions of this work
should therefore explore how to best use past frames and inputs.
This could be coupled with a physics based approach, either as 
part of a controller \cite{yuan2021simpoe} or using explicit physical losses
\cite{Xie_2021_ICCV,Zhang:ICCV:2021,rempe2021humor} in $\internalloss$.
Another interesting direction is the use of more effective parameterizations
for the per-step weights \cite{phasenn,gridnn}. 
While all the problems we tackle here are under-constrained and could thus
have multiple solutions, the current system returns only one.
Therefore, combining the proposed system with multi-modal regressors
\cite{biggs20203d,kolotouros2021prohmr}
is another possible extension.

\section{Conclusion}
\label{sec:conclusion}

In this work, we propose a learned parameter update rule
inspired from classic optimization algorithms
that outperforms the pure network update and is competitive with
standard optimization baselines. We demonstrate the utility of our algorithm
on three different problem sets,
estimating the \threeD body from \twoD keypoints,
from sparse \hmd signals
and fitting the face to dense \twoD landmarks.
Learned optimizers combine the advantages of classic optimization
and regression approaches. They greatly simplify the development process
for new problems, since the parameter priors are directly learned from the data,
without manual specification and tuning, and they run at interactive speeds,
thanks to the development of specialized software for neural network inference.
Thus, we believe that our proposed optimizer will be useful for any applications
that involve generative model fitting.

{%
\small
\qheading{Acknowledgement:} We thank
Pashmina Cameron, Sadegh Aliakbarian,
Tom Cashman, Darren Cosker
and Andrew Fitzgibbon
for valuable discussions and proof reading.
}

\clearpage

{\small
  \balance
  \bibliographystyle{splncs04}
  \bibliography{egbib,amass}
}

\clearpage







\makeatletter
\newcommand{\@chapapp}{\relax}%
\makeatother

\appendixpageoff
\begin{appendices}
    \renewcommand{\thefigure}{A.\arabic{figure}}
    \setcounter{figure}{0}
    \renewcommand{\thetable}{A.\arabic{table}}
    \setcounter{table}{0}

    \chapter*{\supmatlong} \label{App: AppendixA}

\section{Social impact}
Accurate tracking is a necessary pre-requisite for the
next generation of communication and entertainment through
virtual and augment reality. Learned optimizers represent a promising
avenue to realize this potential. However, it can also be used
for surveillance and tracking of private activities of an individual,
if the corresponding sensor is compromised.

    \begin{figure}[t]
\begin{center}
   \includegraphics[trim=000mm 000mm 000mm 000mm, clip=true, width=1.00\linewidth]{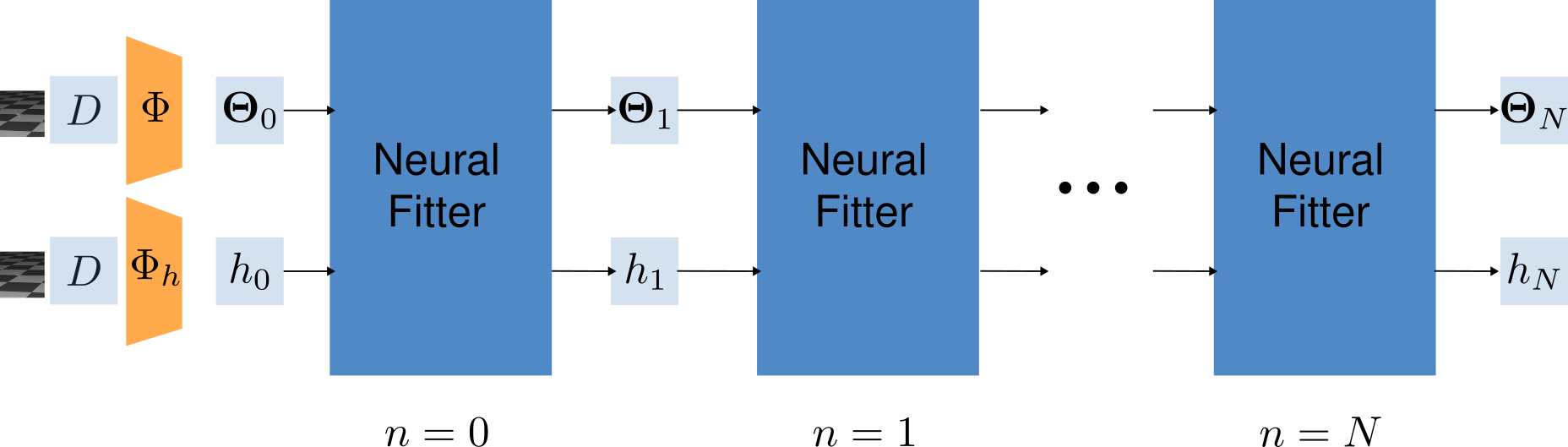}
   \includegraphics[trim=000mm 000mm 000mm 000mm, clip=true, width=1.00\linewidth]{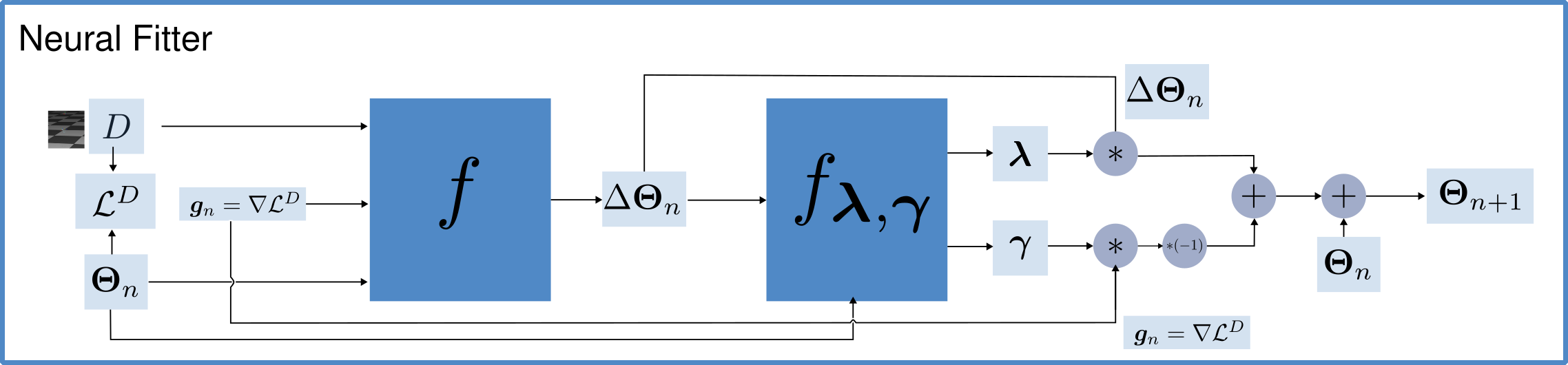}
   \end{center}
    \caption{%
    Top: the general fitting process described in Alg. 1.
    Bottom: A schematic representation of our update rule, 
    described in Eq. 1, 2 of the main paper.
    }
    \label{fig:arch}
\end{figure}
    \begin{figure}[t]
    \begin{center}
        \includegraphics[trim=000mm 000mm 000mm 000mm, clip=true, width=0.33\linewidth]{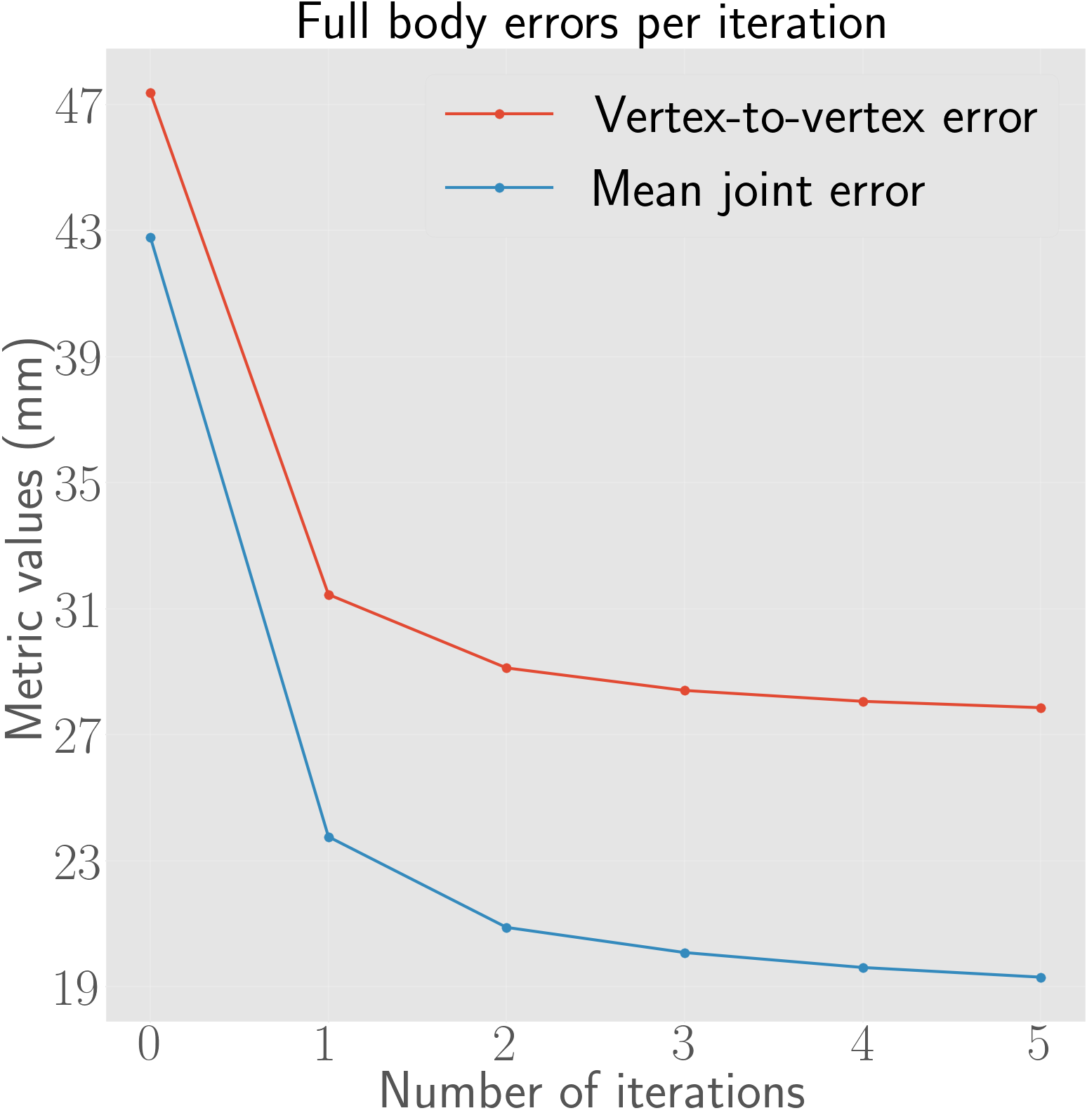}%
        \includegraphics[trim=000mm 000mm 000mm 000mm, clip=true, width=0.33\linewidth]{images/full_error_per_iter/parts.pdf}%
        \includegraphics[trim=000mm 000mm 000mm 000mm, clip=true, width=0.33\linewidth]{images/full_error_per_iter/gnd_penetration.pdf}
    \end{center}
    \caption{%
        Errors per iteration when fitting \smplh
        to \hmd data, assuming that the hands are always visible.
        From left to right: 1) Full body vertex and joint errors, 2) head, left and
        right hand \vtov errors and 3) vertex and joint ground distance,
        computed on the set of points below ground.
    }
    \label{fig:error_per_iter_full_vis}
\end{figure}
    \section{Errors per iteration}

\Fig{fig:error_per_iter_full_vis} shows the metric
values per iteration, averaged across the test set,
for our fitter on the task of fitting \smplh to
\hmd head and hand signals. Different to the main paper,
this figure corresponds to the full visibility scenario,
\ie the hands are always visible. The learned fitter
aggressively optimizes the target data term and quickly
converges to the minimum.

    \section{Update rule}

In addition to the update rule described in Eq. 1 of the main paper, we investigated
two other alternatives, based on the convex combination of the network update and gradient descent.
The first is a simple re-formulation of Eq. 1, with $\damping \in [0, 1]$, selecting
either the network update or the gradient descent direction. 
In the second, we first compute a convex combination between the normalized
network update and gradient descent, \ie selecting a direction,
and then scale the computed direction according to $\lr$.

\begin{equation}
\begin{aligned}
    & u(\Delta\params_n, \grad_n, \params_n) = \damping \Delta\params_n + (1-\damping ) \left(-\lr \grad_n\right) \\
    & u(\Delta\params_n, \grad_n, \params_n) = \lr \left[ \damping \left(\frac{\Delta\params_n}{\lVert \Delta\params_n \rVert} \right) + (1-\damping) \left(\frac{-\grad_n}{\lVert \grad_n \rVert}  \right)
    \right]
    \\
   & \damping = \sigma\left(
    f_{\lambda}(\residuals(\params_n),
    \residuals(\params_n + \Delta\params_n))
    , \damping \in \reals^{\cardinality{\params}}
  \right)
\end{aligned}
\end{equation}

Here, $\sigma()$ is the sigmoid function: $\sigma(x) = \frac{1}{1 + \exp{(-x)}}$.
The learning rate of the gradient descent term is the same as the main text:
\begin{equation}
    \lr =
     f_\gamma(\residuals(\params_n), \residuals(\params_n + \Delta\params_n))
    , \lr \in \reals^{\cardinality{\params}} 
\end{equation}

We empirically found that the performance of these two variants is inferior to the
proposed update rule, but we nevertheless list them for completeness.
    \section{Additional ablation}

\Cref{tab:body_twod_ablation} contains an additional ablation experiment,
where we compare different options for the type of variable for $\damping$, 
$\lr$, namely whether to use a scalar or a vector variable, and
and whether to use a common network predictor for $\damping$, 
$\lr$.
We use the problem of fitting \smpl to  \twoD keypoint predictions,
evaluating our results using the \threedpw test set.

\begin{table}[t]
\captionsetup{font=footnotesize}
    \caption{%
    Predicting vector values for $\damping,\lr$ is always better
    than scalars. This is expected, since each variable to be optimized
    has different scale and the learned fitter must adapt its predicted updates
    accordingly. Having a shared network for $\damping,\lr$ improves
    performance and lowers the number of parameters of the learned fitter.
    }
    \renewcommand{\arraystretch}{\myarraystretch}
    \setlength{\tabcolsep}{12pt}
\centering
\resizebox{1.00\linewidth}{!}{
    \begin{tabular}{cccc}
    Vector $\damping$ & Vector $\lr$ & Shared network for $\damping, \lr$ & \pampjpe (mm) \\ 
    \toprule 
    \cmark & \xmark & \xmark & 52.8 \\
    \xmark & \cmark & \xmark & 52.7 \\
     \cmark & \cmark & \xmark & 52.3 \\
     \cmark & \cmark & \cmark & \textbf{52.2} \\
    \bottomrule
    \end{tabular}
    }
    
    \label{tab:body_twod_ablation}
\end{table}

    \section{Qualitative comparisons}

\begin{figure*}[!t]
\centering

 \includegraphics[trim=200mm 000mm 260mm 020mm, clip=true, width=0.20\linewidth]{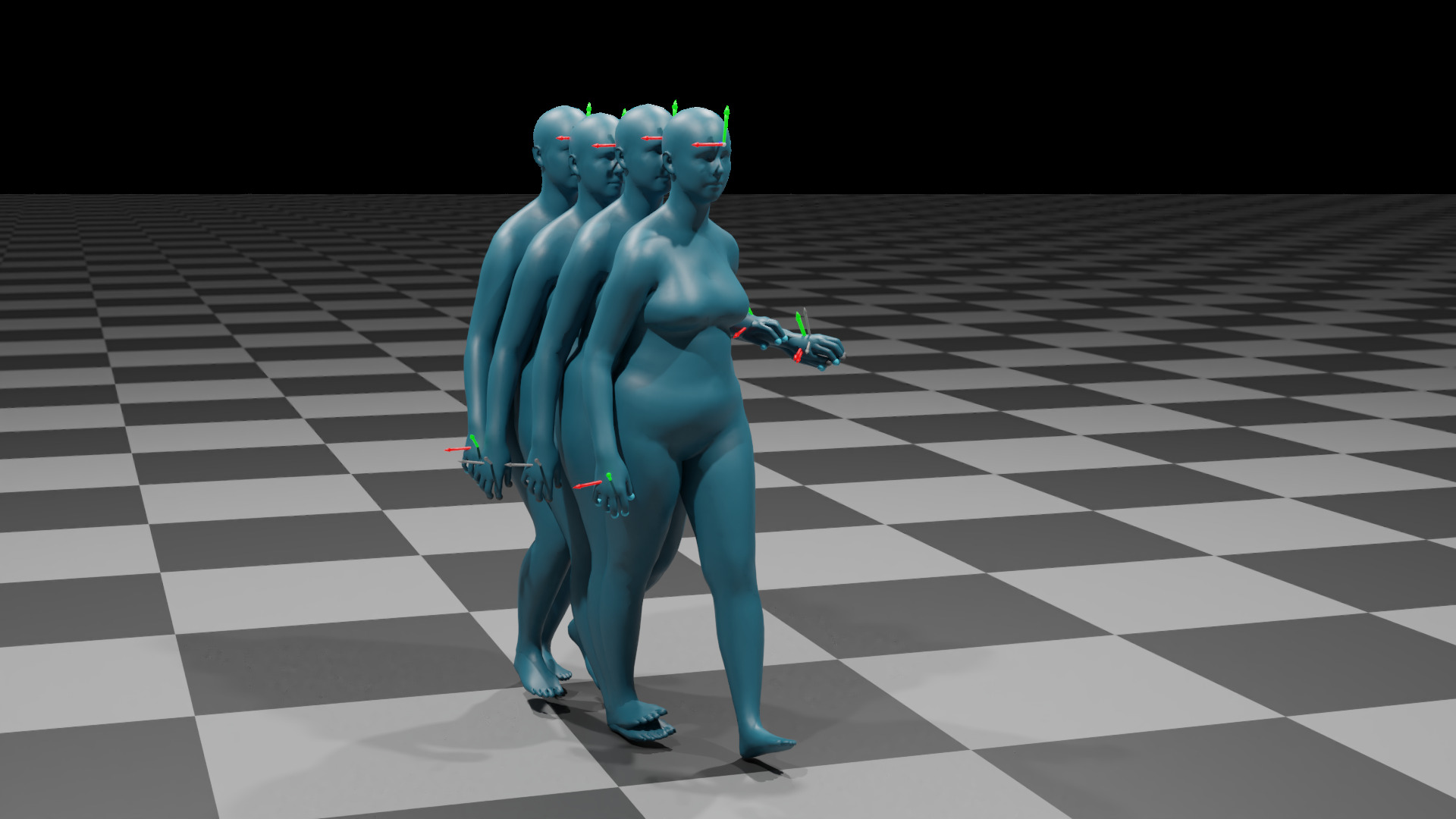}
 \includegraphics[trim=200mm 000mm 260mm 020mm, clip=true, width=0.20\linewidth]{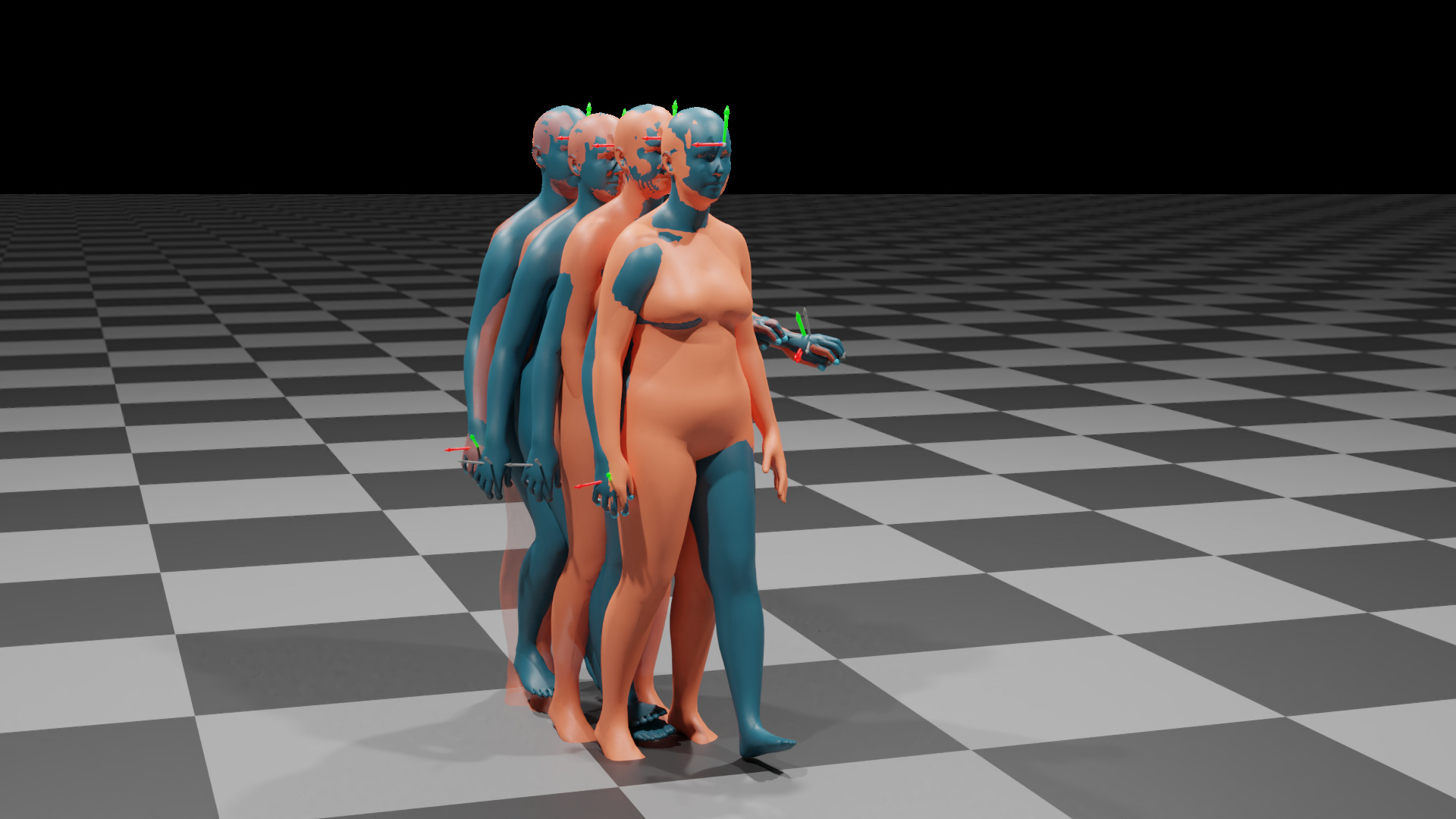}
 \includegraphics[trim=200mm 000mm 260mm 020mm, clip=true, width=0.20\linewidth]{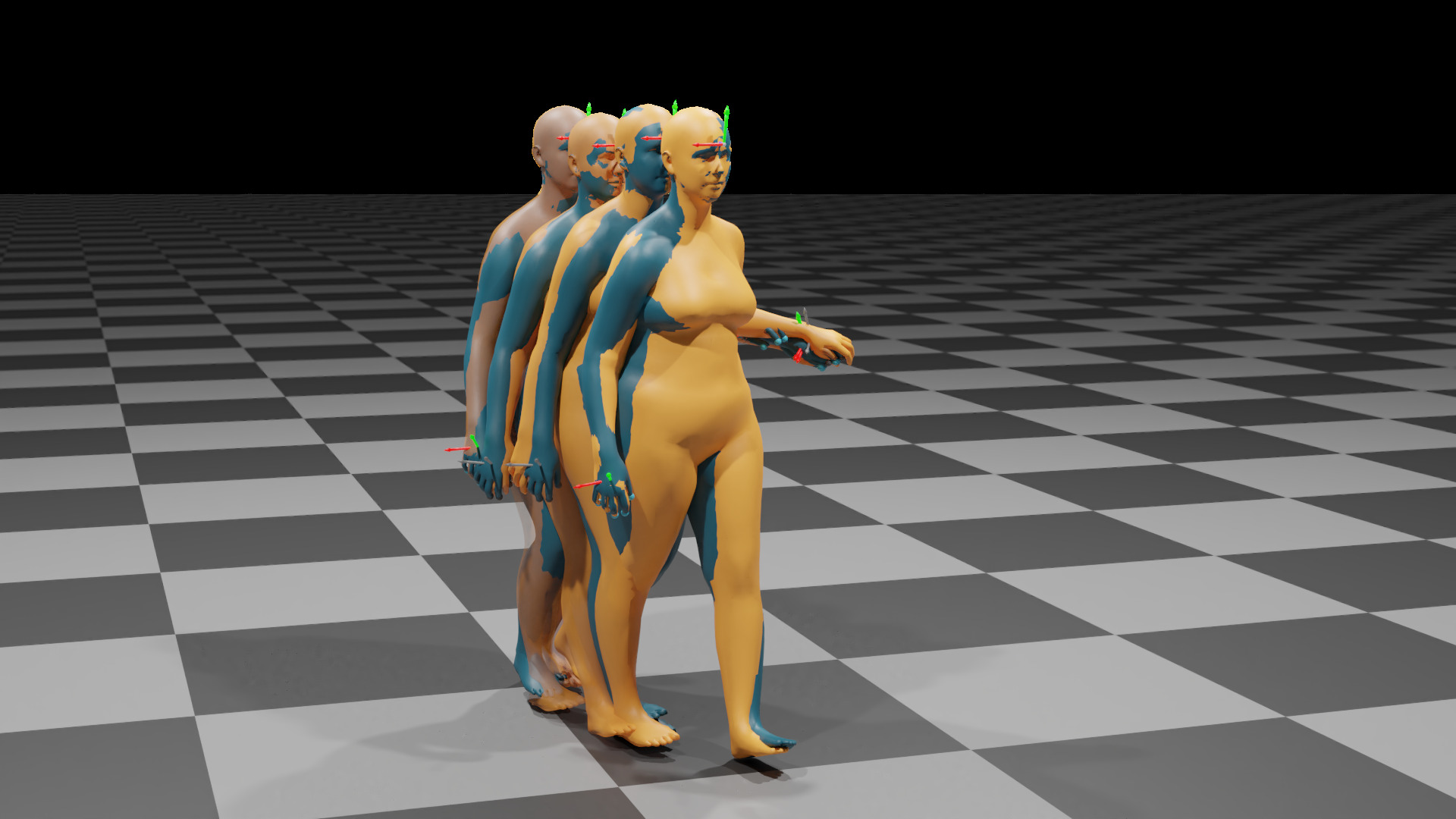}

 \includegraphics[trim=280mm 080mm 220mm 060mm, clip=true, width=0.20\linewidth]{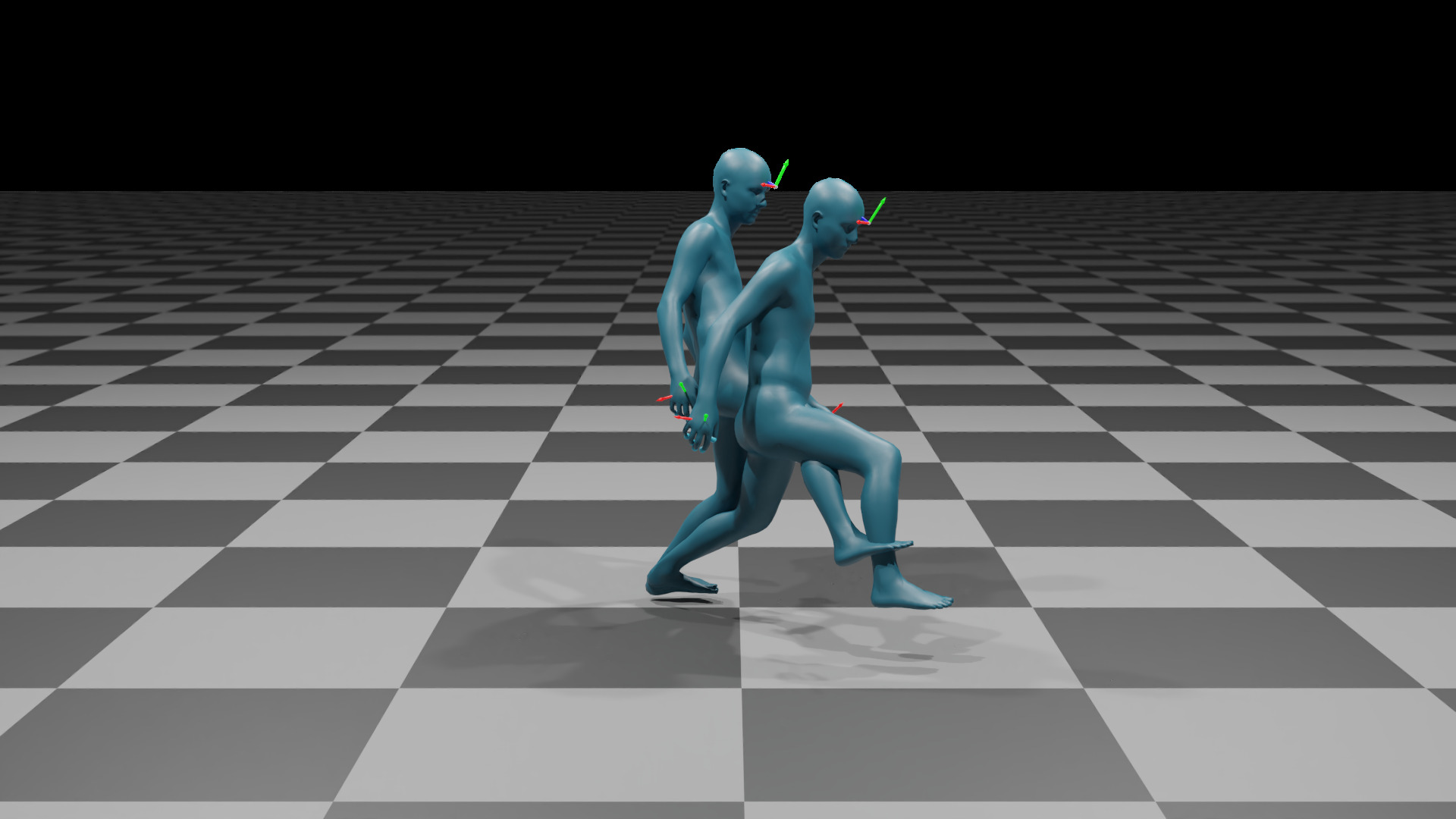}%
 \includegraphics[trim=280mm 080mm 220mm 060mm, clip=true, width=0.20\linewidth]{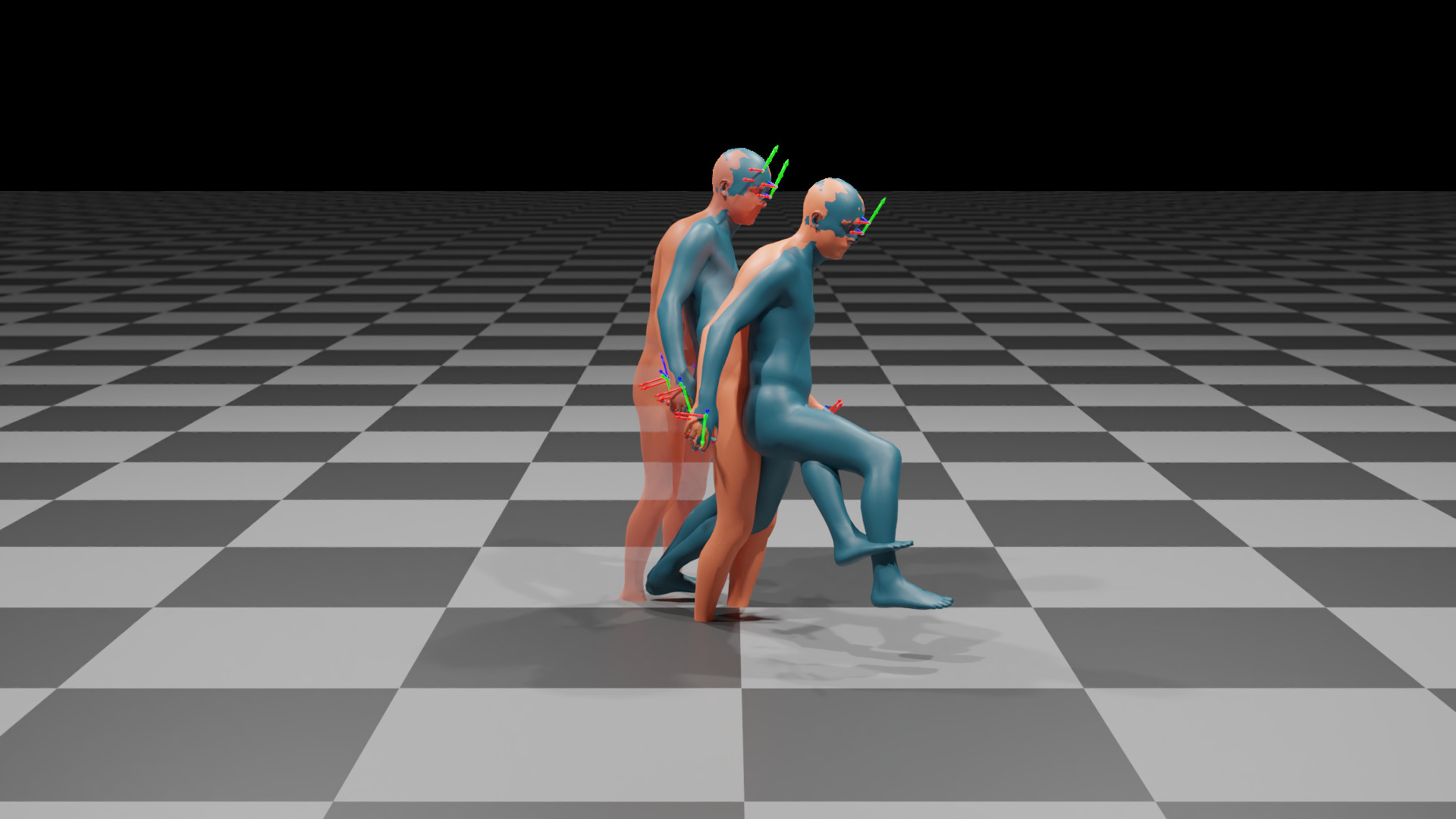}%
 \includegraphics[trim=280mm 080mm 220mm 060mm, clip=true, width=0.20\linewidth]{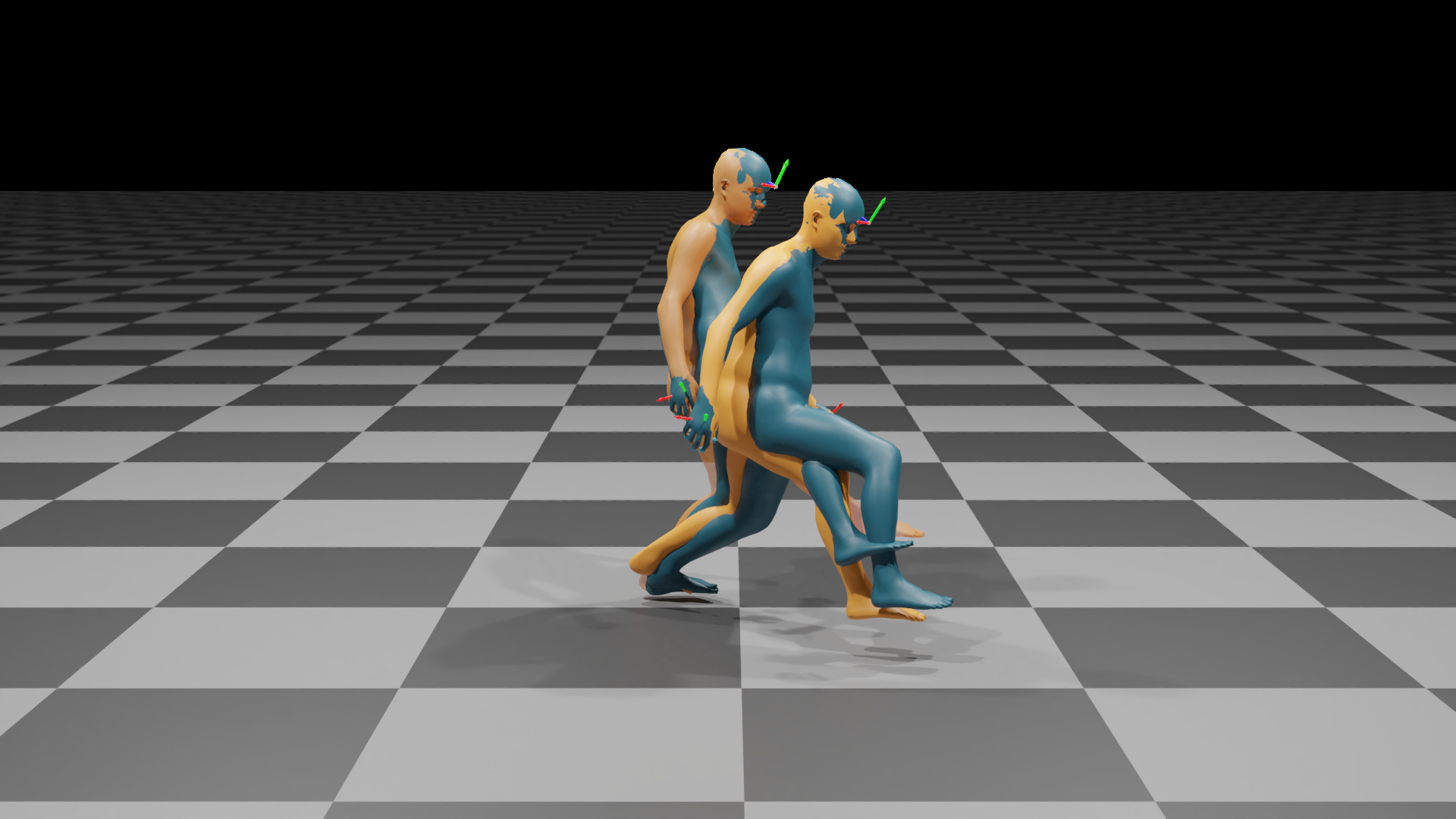}

\includegraphics[trim=280mm 100mm 260mm 060mm, clip=true, width=0.20\linewidth]{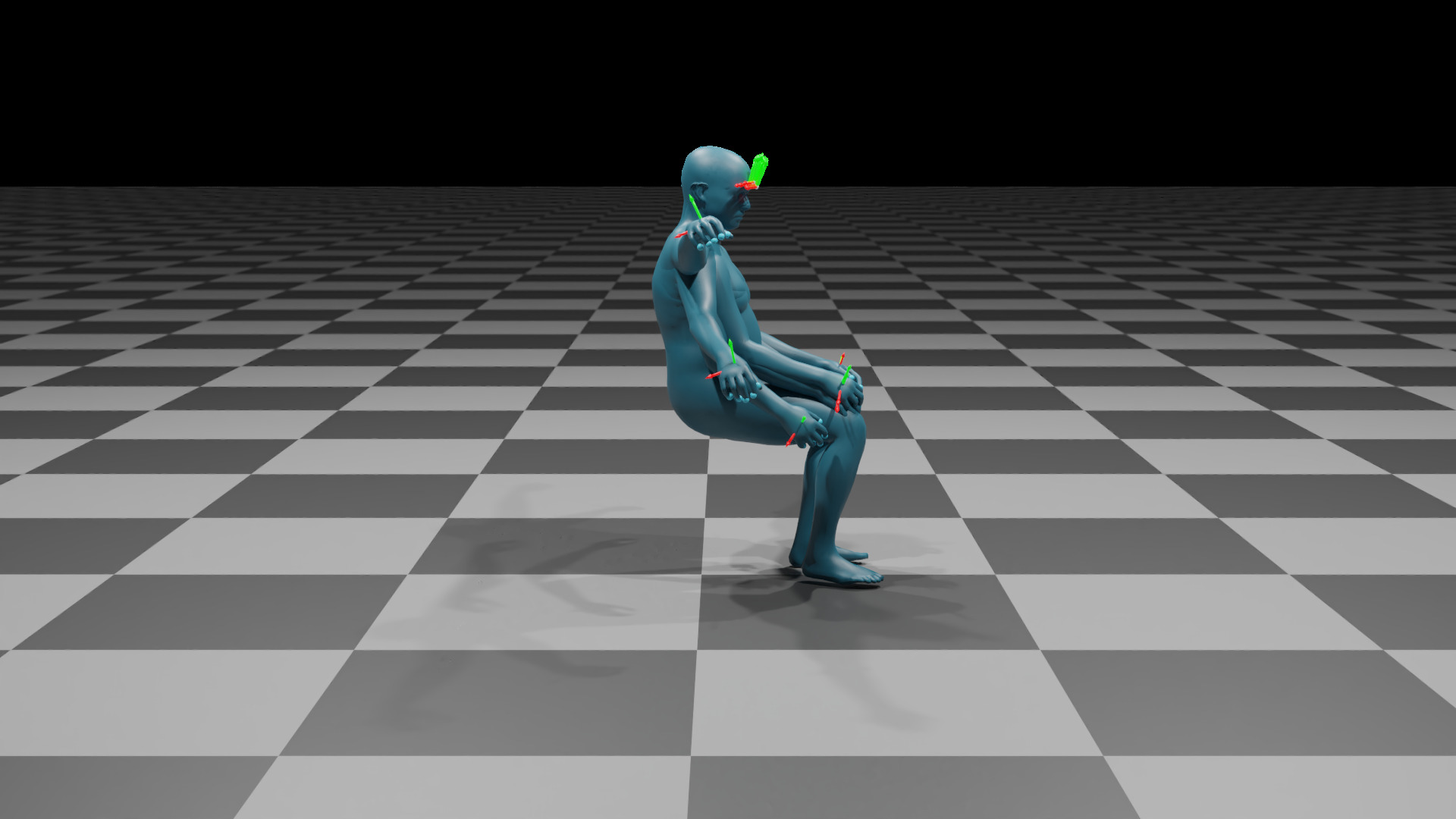}
\includegraphics[trim=280mm 100mm 260mm 060mm, clip=true, width=0.20\linewidth]{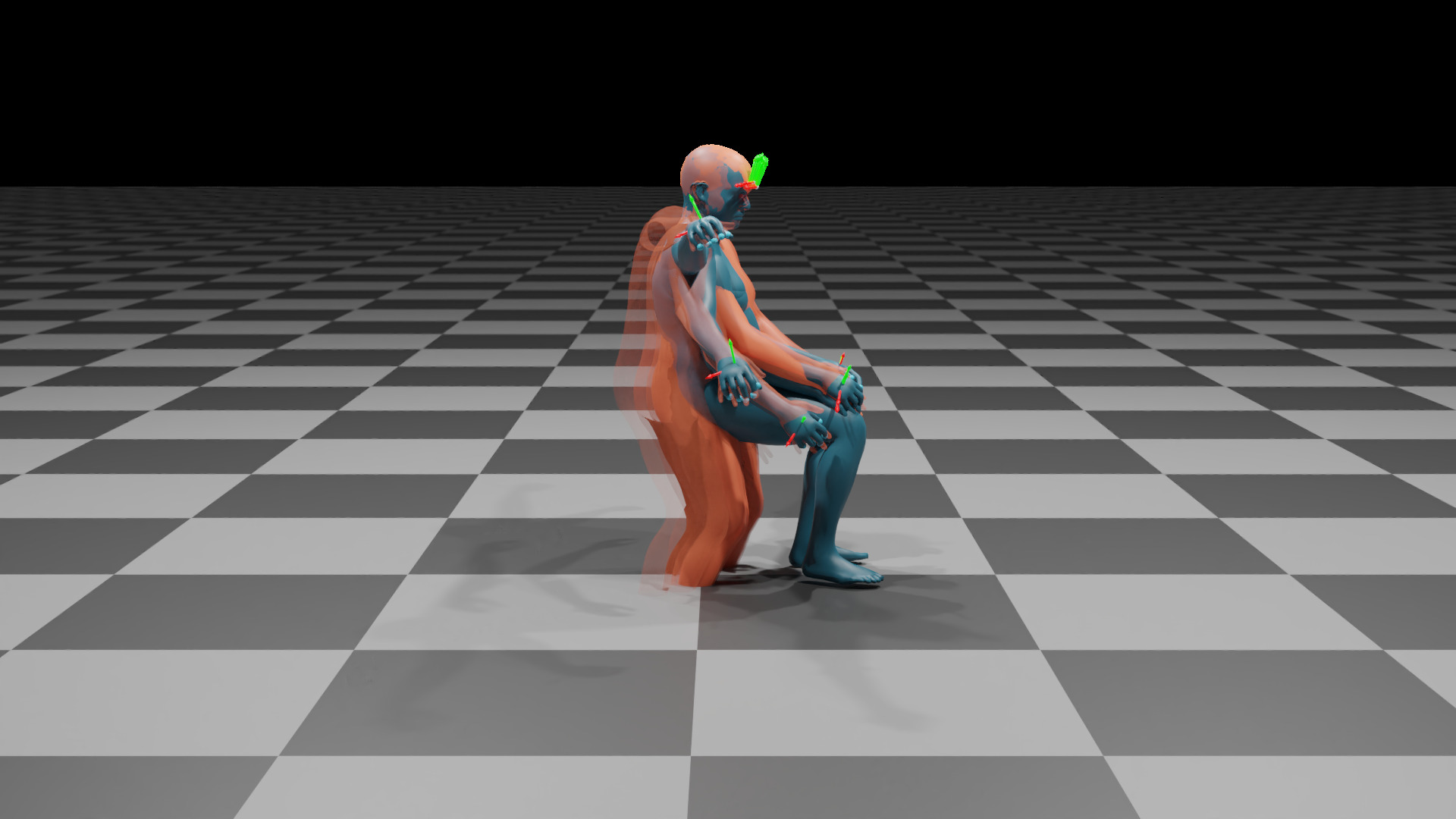}
\includegraphics[trim=280mm 100mm 260mm 060mm, clip=true, width=0.20\linewidth]{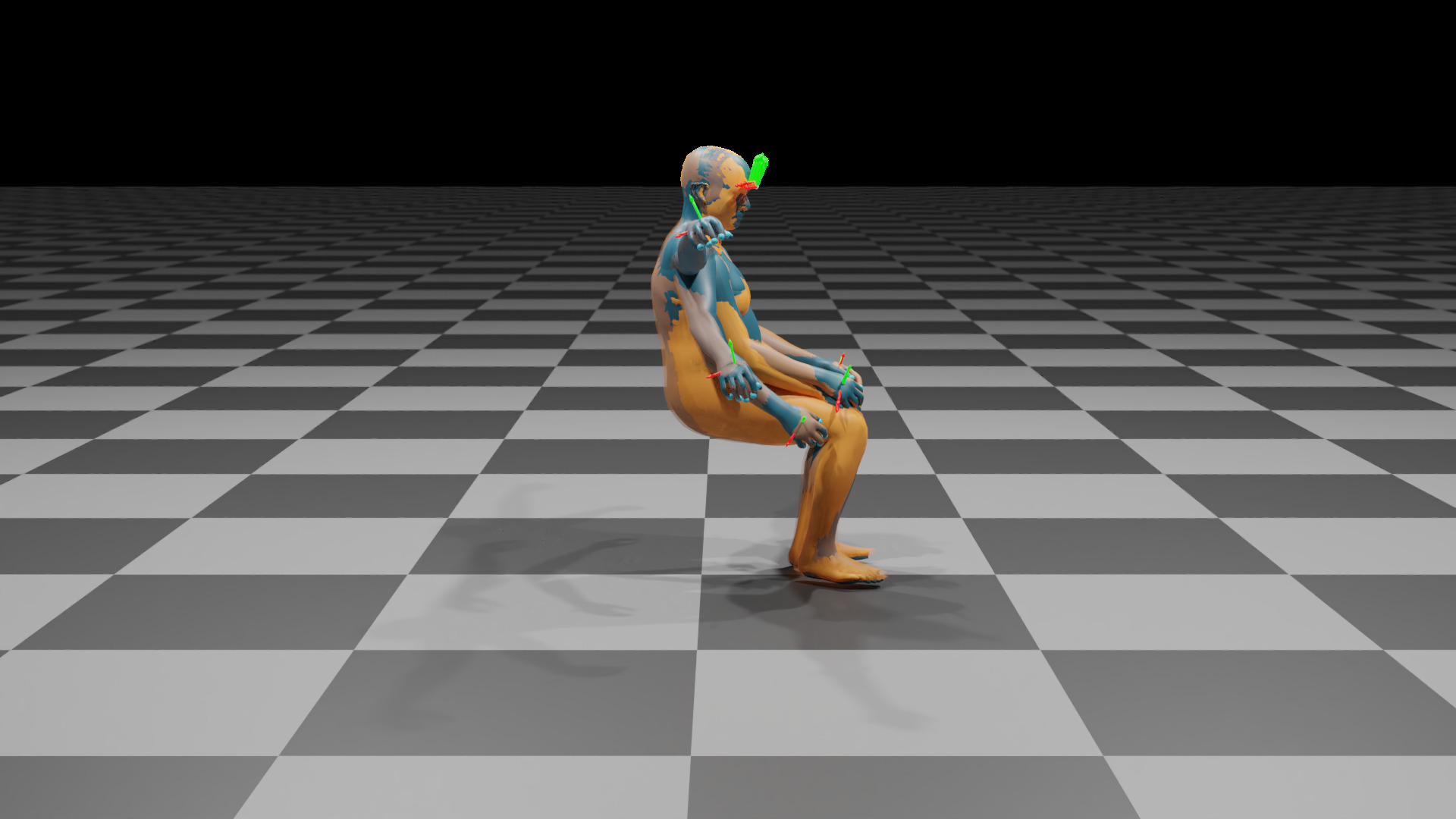}

\includegraphics[trim=280mm 100mm 260mm 060mm, clip=true, width=0.20\linewidth]{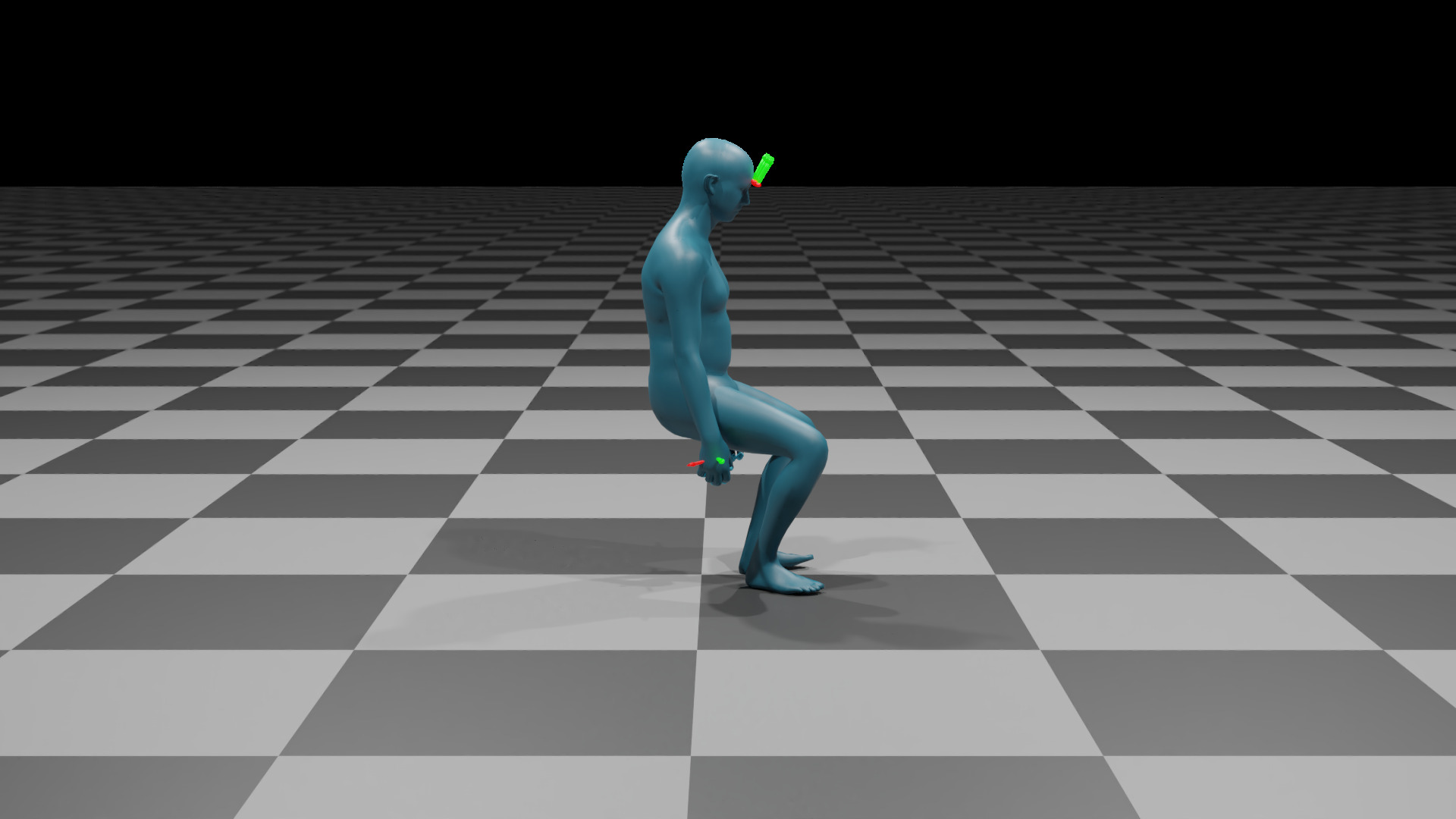}
\includegraphics[trim=280mm 100mm 260mm 060mm, clip=true, width=0.20\linewidth]{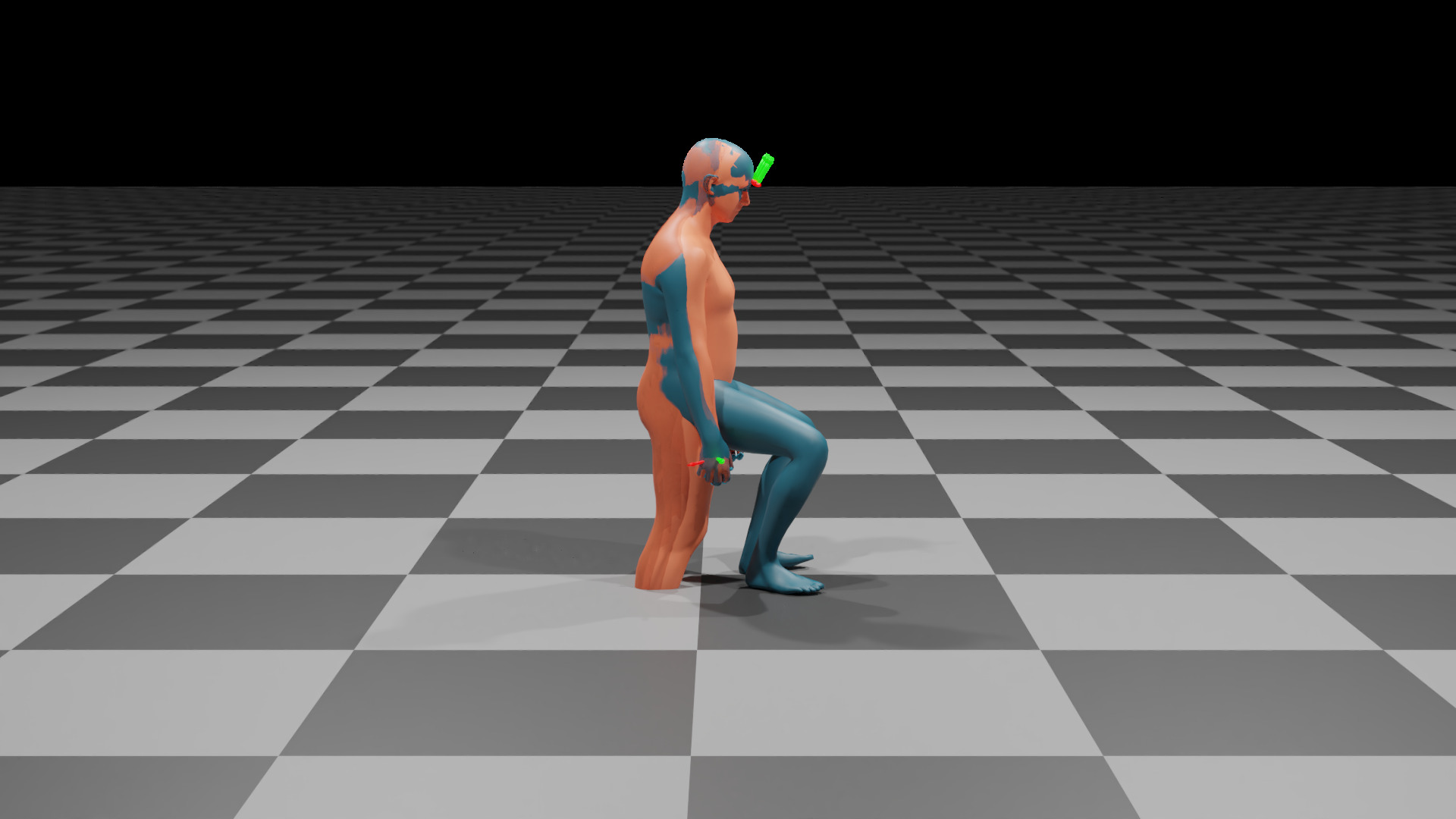}
\includegraphics[trim=280mm 100mm 260mm 060mm, clip=true, width=0.20\linewidth]{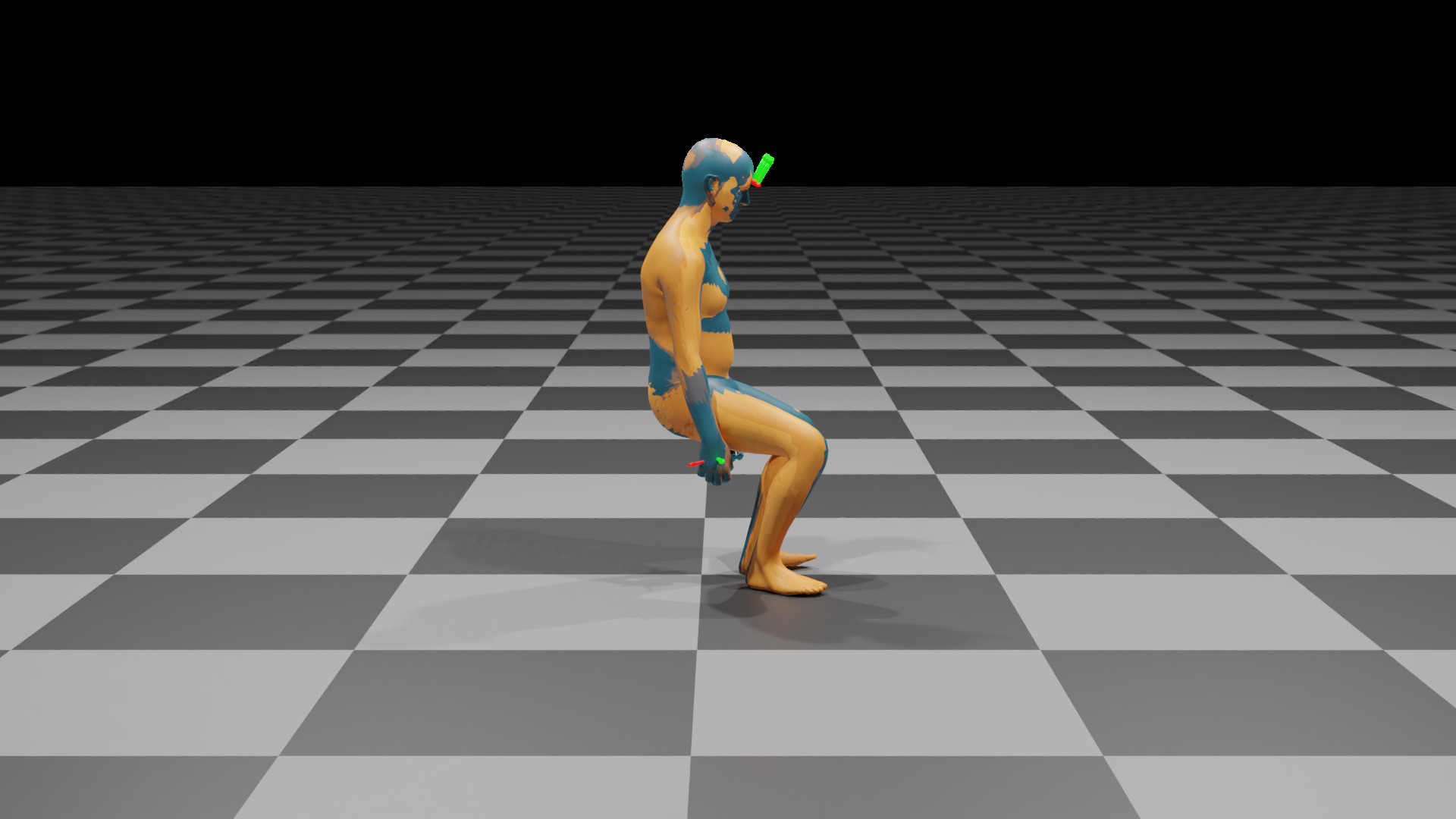}

\caption{%
Comparison of our learned fitter
with a Levenberg-Marquardt based optimization method.
Left to right: 1) Input \hmd data and Ground-Truth
mesh (blue), 2) LM solution (orange) overlayed on the GT, 3)
our solution (yellow) overlayed on the GT.
While the classic LM optimization successfully fits the input data,
it still needs hand-crafted priors to prevent
ground floor penetration. In contrast, our proposed
fitter learns from the data
to avoid such penetrations.
}
\label{fig:hmd_lm_qualitative}

\end{figure*}

We present a qualitative comparison of the proposed
learned optimizer
with a classic optimization-based method
in \fig{fig:hmd_lm_qualitative}.
Without explicit hand-crafted constraints, the classic
approach cannot resolve problems such as ground-floor
penetration. Formulating a term to represent this constraint
is not a trivial process.
Furthermore, tuning the relative weight of this term
to avoid under-fitting the data term is not a trivial
process. Our proposed method on the other hand can
learn to handle these constraints directly from data,
without any heuristics.

    \section{Training details}

\subsection{\gru formulation}

All our recurrent networks are implemented with \grulong{s} (\gru) \cite{gru},
with layer normalization \cite{ba2016layer}:
\begin{equation}
    \begin{aligned}
    & z_n = \sigma_g \left(LN(W_z x) + LN(U_z h_{n-1}) \right) \\
    & r_n = \sigma_g \left(LN(W_r x) + LN(U_r h_{n-1}) \right) \\
    & \hat{h}_n = \phi_h \left(LN(W_h x) + LN(U_h \left(r_n \odot h_{n-1}\right)) \right) \\
    & h_n = (1 - z_n) \odot h_{n-1} + z_n \odot \hat{h}_{n}, 
    \quad h_0 = \Phi_h \left(\data \right)
    \end{aligned}
    \label{eq:gru} %
\end{equation}

We also tried replacing the \gru{s} with LSTMs \cite{hochreiter1997long}, but
did not observe significant performance benefit. Hence we chose the computationally lighter 
\gru{s}.

\subsection{Training losses}

We apply a loss on the output
of every step of our network:
\begin{equation}
    \loss(
    \{\params_n\}_{n=0}^N,
    \{\hat{\params}_n\}_{n=0}^N ;
    \data
    ) = \sum_{i=0}^N \loss_i(\params_i, \hat{\params}_i; \data)
    \label{eq:loss_sum}
\end{equation}
The loss $\loss_i$ contains the following terms:
\begin{align}
    &\loss_i = \lambda_\mesh \loss_i^{\mesh} + 
    \lambda_{\edges} \loss_i^{\edges} + 
    \lambda_{T}\loss_i^{T} + 
    \lambda_{\pose}\loss_i^{\pose} 
    \\
    &\loss_i^{\mesh}  = \normabs{\hat{\mesh}-\mesh} \\
    &\loss_i^{\edges}  = \sum_{(i, j) \in \edges}
    \normabs{%
    (\hat{M}_i - \hat{M}_j) - (M_i - M_j) }
    \\
    &\loss_i^{T} = \sum_{j=1}^J \normabs{\hat{T}_j - T_j} \\
    &\loss_i^{\pose} = \normabs{\hat{\rotation}_{\pose} - \rotation_{\pose}}
    + \normabs{\hat{\transl}-\transl}
\end{align}

$M$ represents the
mesh vertices deformed by parameters $\params$.
$\edges$ is the set of vertex indices of the mesh edges. $T$ denotes the transformations in world coordinate while $\rotation_{\pose}$ denotes the rotation matrices
(in the parent-relative coordinate frame) computed from
the pose values $\pose$. $\transl$ is the root translation vector. 
We use the following values for the weights of the training
losses: 
$\lambda_{\mesh} = 1000$,
$\lambda_{\edges} = 1000$,
$\lambda_{T} = 100$, 
$\lambda_{\pose} = 1$,
$\lambda_{\transl} = 100$.

\subsection{Datasets}

For body fitting from HMD signals, we use a subset of
\amass \cite{AMASS:ICCV:2019} to train and test our method.
Specifically, we use \cmu \cite{cmuWEB}, \kit \cite{KIT_Dataset} and \mpihdm \cite{MPI_HDM05}, adopting 
the same pre-processing and
training, test splits as \cite{dittadi2021full-body}.
An important difference is that we fit the neutral \smplh
to the gendered \smplh data found in \amass, to preserve
correct contact with the ground and avoid the use of heuristics \cite{rempe2021humor}.
We attach random hand poses from the \mano \cite{mano} 
training set to simulate hand articulation.
In all our experiments that involve \smplh, we use
the ground-truth shape parameters $\shape$. Future
work could include estimating a subset of the shape parameters
corresponding to height from the position of the headset.
\noindent For the learned fitter that estimates body parameters
from \twoD joints, we use the data, augmentation and evaluation protocol
of Song \etal \cite{jie_eccv_2020}.
To be more precise, we use \amass \cite{AMASS:ICCV:2019} to train the fitter
and evaluate the resulting model on \threedpw \cite{vonMarcard2018},
which contains sequences
of subjects in complex poses in outdoor scenes, along with \smpl parameters
captured using RGB cameras and IMUs. 

\noindent For face fitting from \twoD landmarks, we use the face model proposed in~\cite{Wood_2021_ICCV} to generate
a synthetic face dataset by sampling $50000$ sets of parameters from the model space. For each sample, we vary pose, identity and expression.
We use a perspective camera with focal length $(512, 512)$ and principal point $(256, 256)$ (in pixels) to project the \threeD landmarks onto the image for \twoD landmarks.
Afterwards, we randomly split this by $80/20$ into training and testing sets. 

\subsection{Training schedule}
We implement our model in \cite{pytorch} and
train it with a batch size of
512 on 4 GPUs using \adam \cite{adam}. We anneal
the learning rate by a factor of 0.1 after 400 epochs.
We apply dropout with a probability of $p=0.5$ on the hidden
states of the \gru{s}. We initialize the weights of the output 
linear layer of \eq{eq:gru} with a gain equal to $0.01$
\cite{glorot2010understanding}.

\subsection{Edge loss}

We empirically observed that the loss between the \threeD edges
of the predicted and ground-truth meshes helps
training converge faster.

\subsection{Runtimes}

We measure time on the \twoD
keypoint fitting problem on a Quadro P5000 GPU and with
a batch size of 512 data points. Our extra networks and
update rule add 6 (ms) per iteration to \lgd's \cite{jie_eccv_2020} runtime.
Using a common network for $\lr$ and $\damping$
 reduces this to 4 (ms).

\subsection{Number of iterations}

Similar to \lgd \cite{jie_eccv_2020}, we observe
limited gains beyond 5 iterations. Training with more
iterations, e.g. 10 or 20, leads to similar performance, at the
cost of increased training time. Picking a random number of
iterations during training, e.g. 5 to 20, does not affect the final
result.

    \begin{figure*}[t]
\centering

\includegraphics[trim=000mm 000mm 000mm 000mm, clip=true, width=0.24\linewidth]{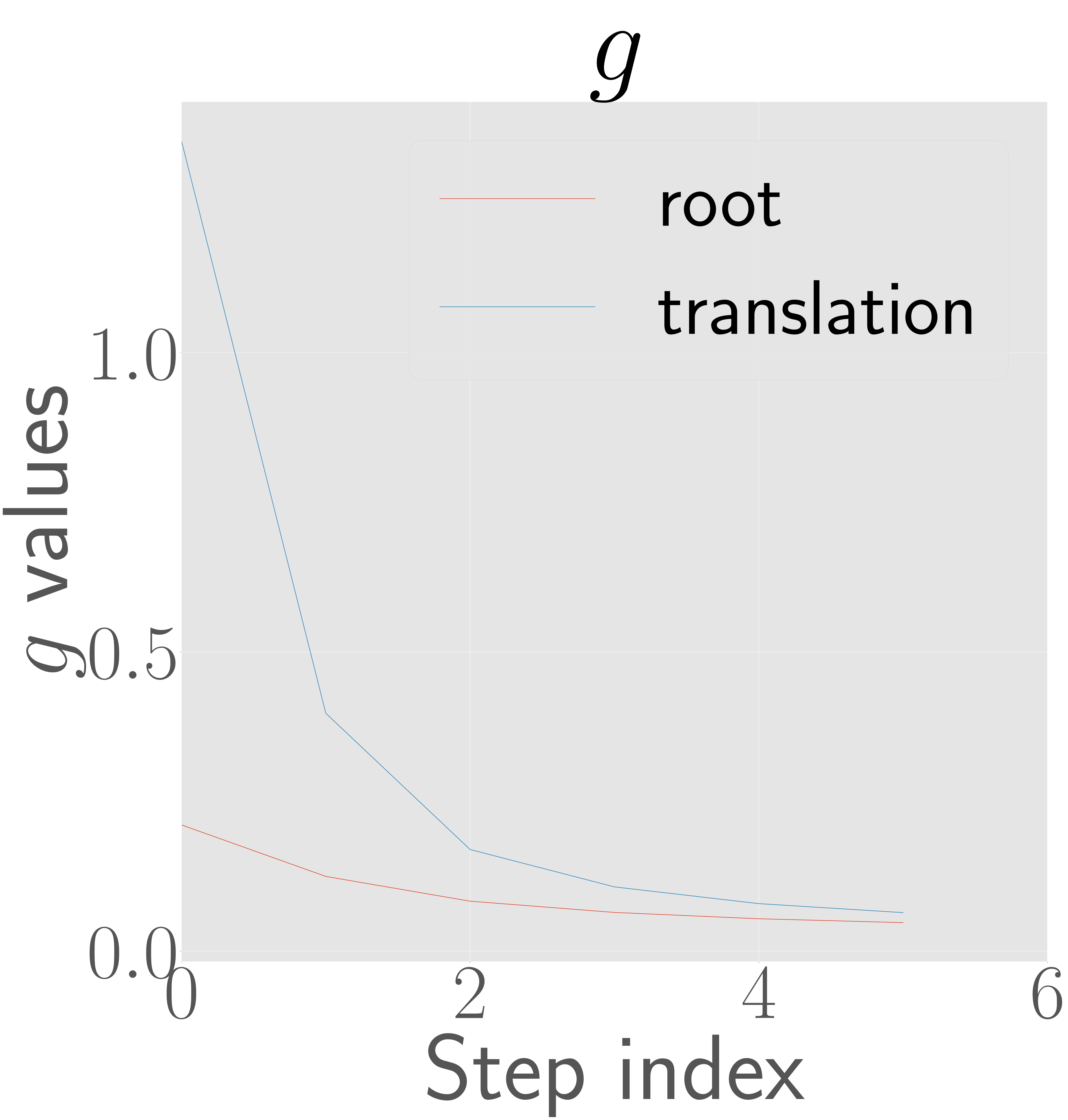}
\includegraphics[trim=000mm 000mm 000mm 000mm, clip=true, width=0.24\linewidth]{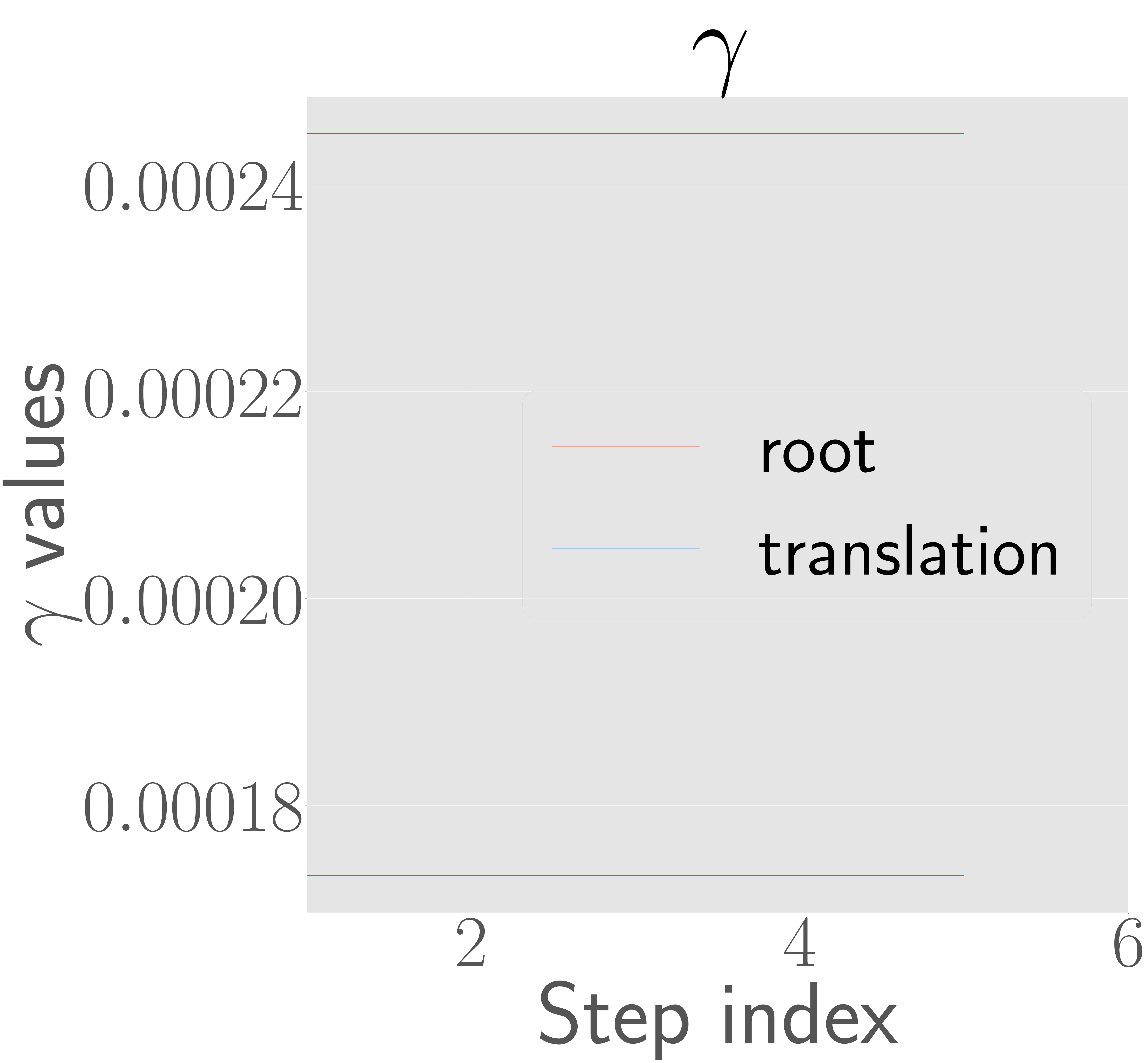}
\includegraphics[trim=000mm 000mm 000mm 000mm, clip=true, width=0.24\linewidth]{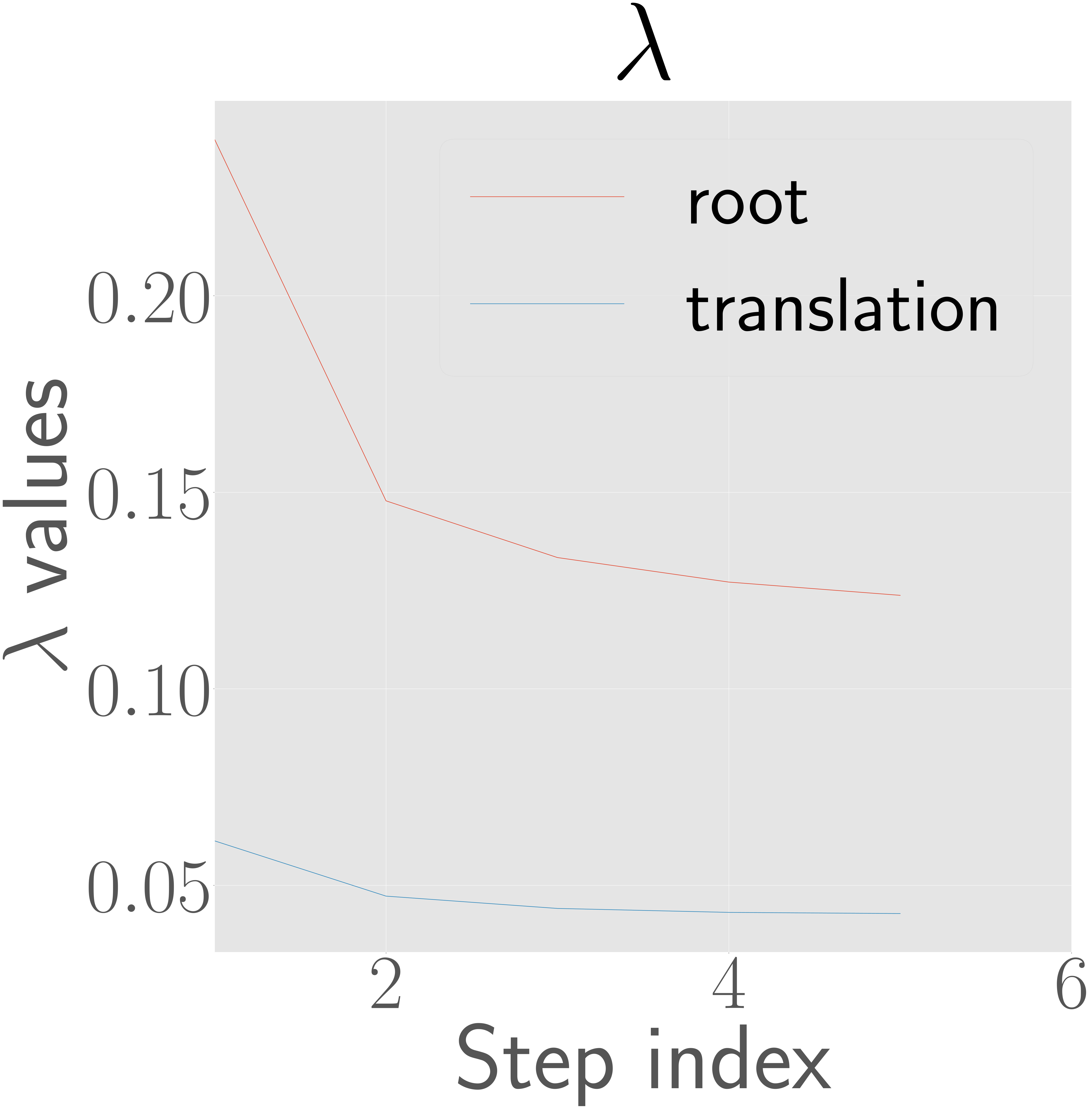}
\includegraphics[trim=000mm 000mm 000mm 000mm, clip=true, width=0.24\linewidth]{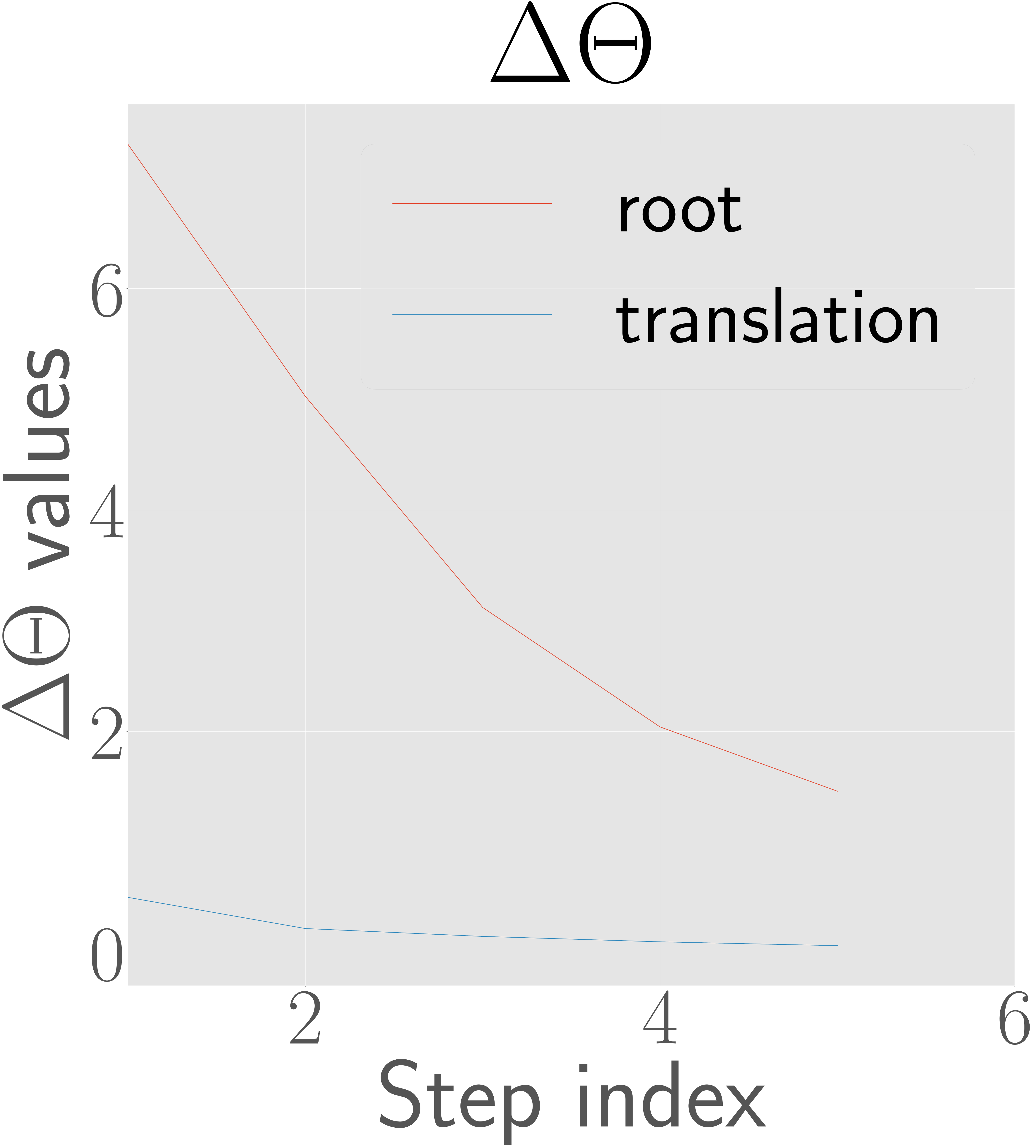}

\caption{%
Average norm for
(left to right) 1) $\normmse{\grad_n}$, 
2) $\normmse{\lr}$,
3) $\normmse{\damping}$
and
4) $\normmse{\Delta\params_n}$,
computed across the test set, for the root rotation and translation.
The learned optimizer slows down as it approaches a minimum of the target
data term.
}

\label{fig:update_values}

\end{figure*}

\end{appendices}




\end{document}